%% file: acl_latex.tex
\pdfoutput=1

\documentclass[11pt]{article}

\usepackage[final]{acl}

\usepackage{times}
\usepackage{latexsym}

\usepackage[T1]{fontenc}

\usepackage[utf8]{inputenc}

\usepackage{microtype}

\usepackage{inconsolata}

\usepackage{graphicx}
\input{tool}

\title{
\ccc: A Bilingual Benchmark for Spoken Dialogue Models Exploring \underline{C}hallenges in \underline{C}omplex \underline{C}onversations}

\author{
  Chengqian Ma\textsuperscript{1}\thanks{Equal contribution.}\thanks{Work is done during internship at LIGHTSPEED.},
  Wei Tao\textsuperscript{2}\footnotemark[1],
  Yiwen Guo\textsuperscript{3}\thanks{Corresponding author.} \\
  \textsuperscript{1}Peking University, 
  \textsuperscript{2}LIGHTSPEED, 
  \textsuperscript{3}Independent Researcher \\
  \texttt{\href{mailto:chengqianma@yeah.net}{chengqianma@yeah.net}}, ~\texttt{\href{mailto:wtao@ieee.org}{wtao@ieee.org}}, ~\texttt{\href{mailto:guoyiwen89@gmail.com}{guoyiwen89@gmail.com}}
}

\begin{document}
\maketitle

\begin{figure}
\vspace{-3em}
\fontsize{12}{6}\selectfont
\begin{tabular}{@{\hspace{6em}}l@{\hspace{6.5em}}c@{\hspace{5.5em}}r@{}}
    \href{https://huggingface.co/datasets/ChengqianMa/C3}{%
        \textmd{Dataset}
    } & 
  \href{https://step-out.github.io/C3-web}{%
    \includegraphics[height=1.8ex]{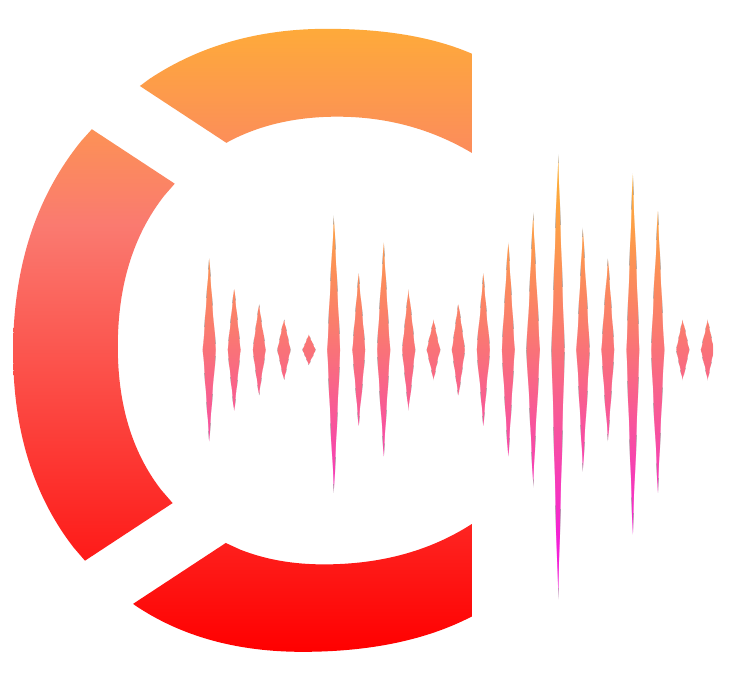}\,\textmd{Website}} &
    \href{https://github.com/step-out/C3}{%
        \textmd{GitHub Repo}
    }
\end{tabular}
\normalsize
\end{figure}

\begin{abstract}
Spoken Dialogue Models (SDMs) have recently attracted significant attention for their ability to generate voice responses directly to users' spoken queries. 
Despite their increasing popularity, there exists a gap in research focused on comprehensively understanding their practical effectiveness in comprehending and emulating human conversations. 
This is especially true compared to text-based Large Language Models (LLMs), which benefit from extensive benchmarking. 
Human voice interactions are inherently more complex than text due to characteristics unique to spoken dialogue. Ambiguity poses one challenge, stemming from semantic factors like polysemy, as well as phonological aspects such as heterograph, heteronyms, and stress patterns. Additionally, context-dependency, like omission, coreference, and multi-turn interaction, adds further complexity to human conversational dynamics.
To illuminate the current state of SDM development and to address these challenges, we present a benchmark dataset in this paper, which comprises 1,079 instances in English and Chinese.
Accompanied by an LLM-based evaluation method that closely aligns with human judgment, this dataset facilitates a comprehensive exploration of the performance of SDMs in tackling these practical challenges.

\end{abstract}

\vspace{1em}
\section{Introduction}

\begin{figure*}[t!]
  \centering
  \includegraphics[width=1.0\textwidth]{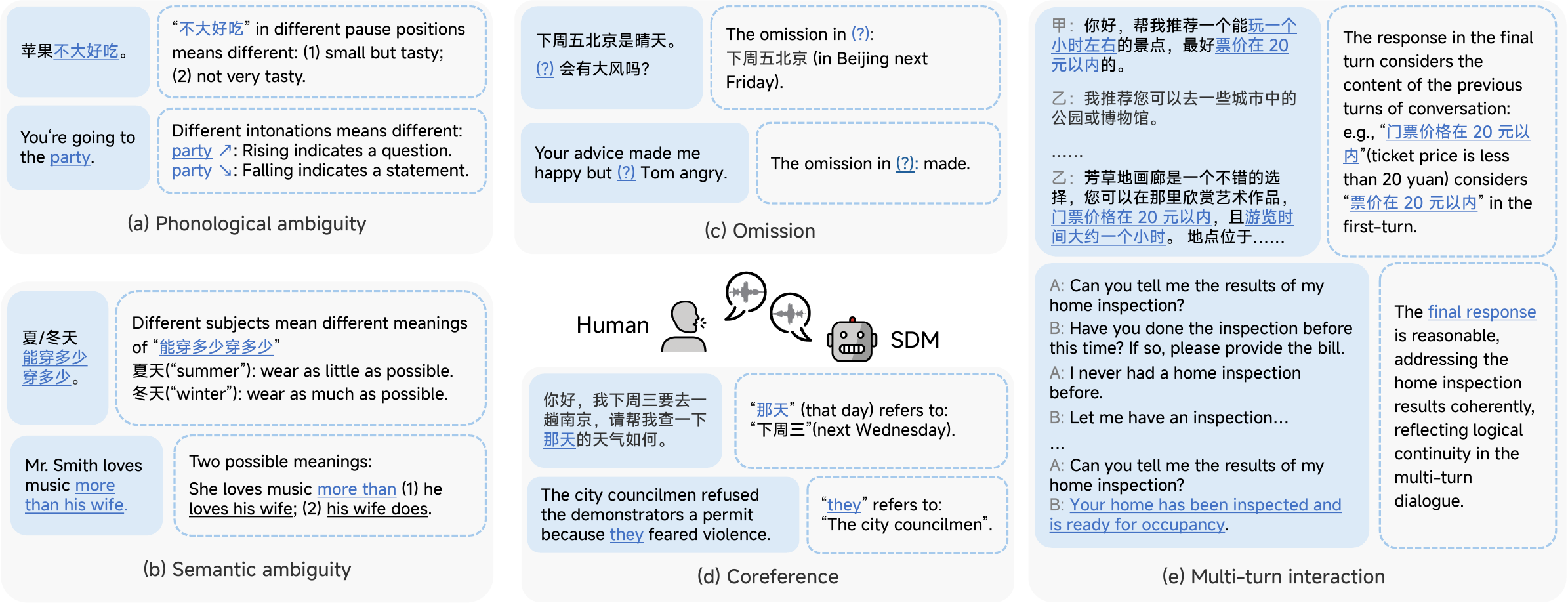}
  \caption{The structure and exemplars within the dataset. The subplots correspond to the sub-datasets of five phenomena. The blue boxes enclose the input for SDM, with some parts of the prompts omitted, while the corresponding outputs are within dashed boxes. Blue underlined text indicates the focal elements of interest, and gray text represents a segment of the prompt. The arrow indicates a rising or falling intonation. The (?) denotes an omitted sentence component. The >> points to the referent of the pronoun. The $...$ represents the omitted dialogue.}
  \label{fig:example_overview}
\end{figure*}

Human conversations, particularly spoken dialogues, are inherently complex owing to ambiguous contexts~\citep{DBLP:journals/corr/SoleS14} that introduce uncertainties in communication. 
Ambiguity arises from phonological elements like pauses and intonation, as well as semantic factors such as lexical and syntactic ambiguity, as demonstrated in Figure~\ref{fig:example_overview}(a) and Figure~\ref{fig:example_overview}(b). 
These ambiguities can lead to misinterpretations, necessitating careful understanding and response from participants. 
Recently, Spoken Dialogue Models (SDMs), such as \gptfouroa~\citep{openai_audio_api} and MooER-Omni~\citep{DBLP:journals/corr/abs-2408-05101}, have become increasingly involved in human interactions. 
An SDM processes voice input and delivers voice response~\citep{DBLP:journals/corr/abs-2411-13577}, and an effective SDM should be capable of recognizing and addressing challenging ambiguities to produce coherent replies.

Even in contexts without ambiguity, challenges can arise for SDMs. 
Speakers may omit previously mentioned entities or those understood as common knowledge, as illustrated in Figure~\ref{fig:example_overview}(c). 
Additionally, speakers often use pronouns to refer to specific entities, as shown in Figure~\ref{fig:example_overview}(d). 
Such context-dependency is significant in multi-turn interaction (Figure~\ref{fig:example_overview}(e)).
This requires SDMs to accurately identify and resolve omissions and coreferences to understand the intent of a speaker. 

Despite the importance of handling ambiguity and context-dependency, it is yet unclear whether current SDMs are capable of addressing these challenges. 
To bridge the gap, we conduct an in-depth empirical study on the complexity of spoken dialogues and propose a novel dataset meticulously designed to study SDMs in handling complex dialogue situations with phonological ambiguity, semantic ambiguity, omission, coreference, and multi-turn interaction. 
Together with the dataset, we also propose an automatic LLM (Large Language Model)-based evaluation method to test the capability of SDMs, which aligns well with human evaluation results.
After studying ten popular SDMs, we deliver three findings to the community, including pointing out the different difficulties of five phenomena, two languages in spoken dialogues, and demonstrating the different advantages of the SDMs.

\section{Related Work}

\subsection{Spoken Dialogue Models}
SDMs can be divided into earlier cascaded models and recent end-to-end models~\citep{DBLP:journals/corr/abs-2411-13577, cui2024recent}. 
The end-to-end model can directly understand and generate speech representations, while the cascaded model consists of Automatic Speech Recognition (ASR)~\citep{DBLP:journals/mta/MalikMMM21, DBLP:conf/icassp/YuCLCSHNHGWP21, DBLP:conf/icassp/HsuTBSM21}, Language Models (LMs), and Text-to-Speech (TTS) modules~\citep{DBLP:conf/icassp/MehtaTBSH24, DBLP:conf/icml/PopovVGSK21}.  
Cascaded models lose crucial audio features (e.g., intonation) during ASR processing, forcing LMs to work only on text. This prevents them from interpreting phonetic phenomena in raw audio.
Consequently, it is natural that they underperform when there exists ambiguity in human speech. 
Our evaluation in this paper thus focuses on end-to-end models.

\gptfouroa~\citep{openai_audio_api} is the first end-to-end SDM that can generate fluent voice responses and analyze the emotions and intonations of the audio input. 
Since the implementation is not public, some open-source works, including LLaMA-Omni~\citep{DBLP:journals/corr/abs-2409-06666} and Freeze-Omni~\citep{DBLP:journals/corr/abs-2411-00774}, are explored and proposed. These works achieve low-latency spoken responses based on LLM in English conversation. 
To achieve real-time full-duplex dialogue capabilities for spoken large language models, Moshi~\citep{DBLP:journals/corr/abs-2410-00037} is proposed, and it supports interruptions.
To support more languages' conversation, MooER-Omni~\citep{DBLP:journals/corr/abs-2408-05101}, GLM-4-Voice~\citep{DBLP:journals/corr/abs-2412-02612}, VITA-Audio~\citep{DBLP:journals/corr/abs-2505-03739}, Step-Audio~\citep{DBLP:journals/corr/abs-2502-11946}, Kimi-Audio~\citep{DBLP:journals/corr/abs-2504-18425}, and Qwen2.5-Omni~\citep{DBLP:journals/corr/abs-2503-20215} are proposed, and they show great ability in both English and Chinese spoken dialogues.
We will study all these mentioned end-to-end SDMs in this paper.

\subsection{Benchmarks and Datasets}
To evaluate the capacities of SDMs, several benchmarks have been developed, each focusing on different aspects of audio~\citep{DBLP:journals/corr/abs-2502-04424,DBLP:journals/corr/abs-2503-21614}. 
ADU-Bench~\citep{DBLP:journals/corr/abs-2412-05167} examines the cross-lingual and cross-skill spoken dialogue understanding capabilities of SDMs. 
Other benchmarks extend beyond language to include additional features. 
For instance, AIR-Bench~\citep{DBLP:conf/acl/YangXLC0ZLLZZZ24} first evaluates the ability to understand various types of audio signals.
SUPERB~\citep{DBLP:conf/interspeech/YangCCLLLLSCLHT21} focuses on speaker and emotion recognition. 
AudioBench~\citep{DBLP:journals/corr/abs-2406-16020} assesses the ability to understand speech, audio scenes, and paralinguistic features. 
SD-Eval~\citep{DBLP:conf/nips/AoWTCZ0W0024} evaluates SDMs' responses to utterances with varying emotions, accents, ages, and background sounds.
MMAU~\citep{DBLP:journals/corr/abs-2410-19168} includes perception and reasoning tasks across speech, sound, and music. 
VoiceBench~\citep{DBLP:journals/corr/abs-2410-17196} focuses on real-world scenarios involving speaker characteristics, environmental conditions, and content factors.

However, these benchmarks have some limitations in four aspects:

(1) Most of the above benchmarks ignore the ambiguity. 
The only exception, ADU-Bench, considers it but does not cover phonological ambiguities such as press, heterograph, heteronym, and some semantic ambiguities, such as syntactic ambiguities.

(2) None of the aforementioned benchmarks consider comprehension difficulties caused by coreference and omission phenomena.

(3) All of the benchmarks listed include real-world spoken dialogue data from only one language (i.e., English). While ADU-Bench incorporates other languages, these datasets are translated from English, which means they may lack language-specific features, such as tone in Chinese. 

(4) These benchmarks focus solely on single-turn dialogues, whereas multi-turn interactions are more common in spoken communication. They do not assess the ability of SDMs to handle multi-turn dialogues.

\section{A New Benchmark for SDMs}

The field of SDMs is rapidly evolving. Few studies could reveal the limitations and real performance of these models in handling complex ambiguity and context-dependency, which widely exist in human conversations.

In this section, we first empirically study each aspect of conversational complexity.
Based on our empirical study, we design the dataset specifically.

\subsection{The Complexity of Spoken Dialogues}\label{sec:motivation}

To investigate the importance of the complex phenomena in spoken dialogue, we conduct a literature review, statistical analysis, and case study.
The statistical analysis is performed using datasets in both English and Chinese. For English dialogues, we use CABank~\citep{macwhinney2010transcribing, yaeger-dror2007cabank_north, yaeger-dror2007cabank_south}. For Chinese dialogues, we use MagicData-RAMC~\citep{DBLP:conf/interspeech/YangCLYYCXJZZX022} as the studied dataset.
These datasets are selected because they are constructed based on real-world spoken dialogues rather than text-based dialogues. 
The reason for not using text-based dialogues is that they differ from spoken dialogues not only in form but also in content~\citep{le2004searching, placinski2023modality}.
Moreover, these two datasets are used in many top conferences~\citep{DBLP:conf/emnlp/GuoYXJX23, DBLP:conf/icassp/LiYYWMMLM21, maheshwari2025asr} and journals~\citep{DBLP:journals/tifs/XieCWY24, DBLP:journals/taslp/LandiniDSB24}.

\subsubsection{Phonological Ambiguity} 
\begin{figure}[htp]
    \centering
    \includegraphics[width=\linewidth]{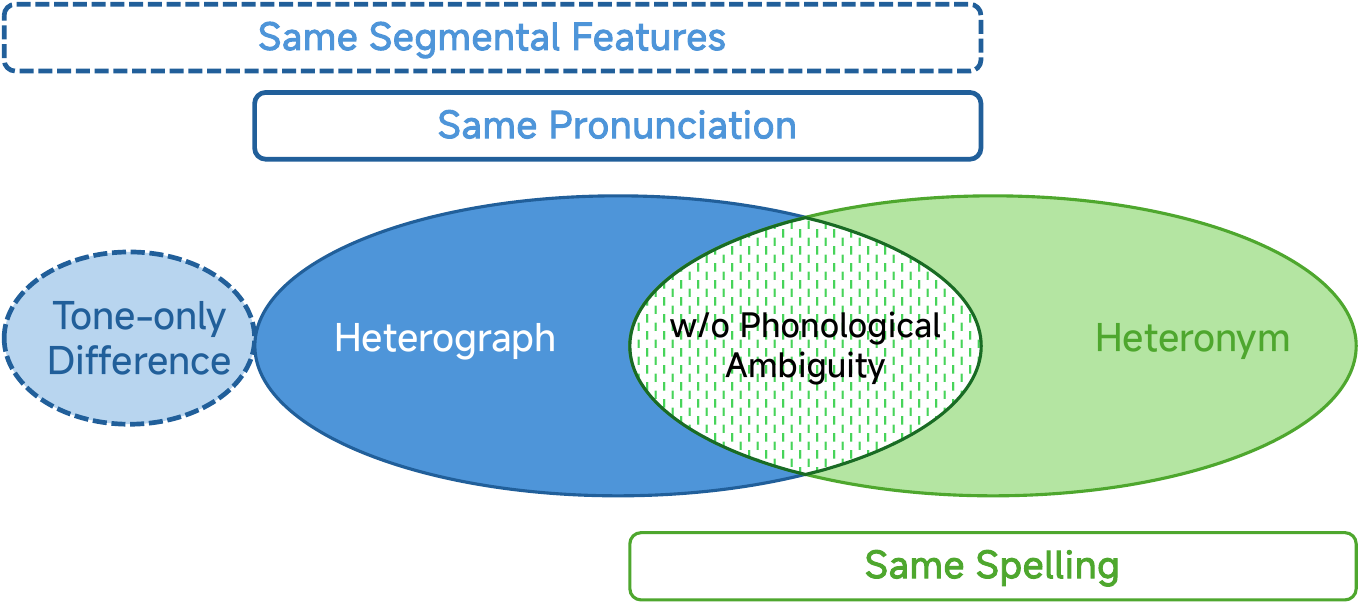}
    
    \caption{The relation between terms in Section 3.1.1.}
    \label{fig:term_pho_amb}
    
\end{figure}

Phonological ambiguity can be classified into two types: segmental and supra-segmental.
The former refers to discrete units that can be identified auditorily in the stream of speech. The latter refers to those features that extend over more than a single unit in an utterance~\citep{ladefoged2006course, sharma2021significance}.
To make this section clearer, some terms are clarified as shown in Figure~\ref{fig:term_pho_amb}. 

Firstly, we investigate the segmental ambiguity. 

\noindent\textbf{Tone-only Difference}: In spoken dialogue, especially in Chinese, the same segmental features do not convey the same meaning. For example, the Chinese phonetic alphabet $hao$ can have four different tones, and each tone refers to a set of Chinese characters. The tone-only difference in pronunciation can lead to ambiguity. We use the tool~\citep{pypinyin} to count the situation in the dataset. We find that more than 99.25\% Chinese characters from real-world dialogues have characters with the same   phonetic alphabet but different tones, which can contribute to the ambiguity.

\noindent\textbf{Heterograph}: Some words with the same pronunciation may have different spellings. For example, in English, ``night'' and ``knight'', ``tail'' and ``tale'' are heterographs\footnote{A word whose pronunciation is the same, but whose spelling and meaning differ from another’s.}. We use the tools~\citep{pypinyin, pronouncing} to count the situation in the dataset and find that there are 7.05\% of the English words and 97.94\% of the Chinese characters in dialogues are heterographs.

\noindent\textbf{Heteronym}: Some words with the same spelling also have different pronunciations. Of the 2,000 most frequently used English words, 9 of them are heteronyms\footnote{A word having the same spelling as another but a different meaning, and often a different pronunciation.}~\citep{parent2012most}. A study~\citep{zhang2002statistical}
reveals that there are at least 688 Chinese heteronyms. We use the tool~\citep{pronouncing} to explore English dialogues and find that at least 851 English heteronyms appear more than 42,315 times in real-world spoken dialogues.

The numbers above demonstrate the widespread existence of each phenomenon that can contribute to the segmental phonological ambiguity.

Secondly, we investigate the supra-segmental ambiguity. 
Pause, intonation, and stress are three supra-segmental features that can lead to ambiguity.
Figure~\ref{fig:example_overview}(a) shows two examples with different pause positions and with different intonations.
The placement of stress in English can lead to ambiguity~\citep{yang2025brief}. 
For example, ``a green house'' refers to a building with a roof and sides made of glass when the word ``green'' is stressed, but it denotes a building that is colored green when the word ``house'' is stressed.

\subsubsection{Semantic Ambiguity}
As shown in Figure~\ref{fig:term_pho_amb}, the words that have the same pronunciation and spelling do not have phonological ambiguity, but semantic ambiguity can exist in them.
Semantic ambiguity can be classified into two types: lexical and syntactic.

\noindent\textbf{Lexical Ambiguity}: It means one word in a sentence can have two or more meanings. For example, in the sentence ``They exchanged addresses in darkness'', the term ``darkness'' can be interpreted as either ``in the absence of light'' or ``secretly''.
A study on 11 business articles~\citep{jannah2021lexical} identified 27 instances of lexical ambiguity, demonstrating the widespread presence.

\noindent\textbf{Syntactic Ambiguity}: This means the situation where a sentence can be interpreted in more than one way due to its grammatical structure. Examples are shown in Figure~\ref{fig:example_overview}(b).
We use the tool~\citep{spacy_usage} to analyze the dataset and find that there are 15.79\% of Chinese and 41.14\% of English sentences with syntactic ambiguity in dialogues.

The numbers mentioned above demonstrate that semantic ambiguity often occurs in spoken dialogues.

\subsubsection{Omission}
Omission (also known as ellipsis) is common in spoken conversations. Two examples are shown in Figure~\ref{fig:example_overview}.
Moreover, subjects, verbs, and pronouns can be omitted in English dialogues~\citep{mcshane2005theory}. Statistically, a study~\citep{glass2022english} finds that the omission of verb objects is particularly common when describing routines. Another study~\citep{DBLP:conf/acl/SuSZSHNZ19} shows that 52.4\% of Chinese utterances also have omissions in dialogues. 

We use the tools~\citep{spacy_usage} for analysis and find that the incidence of subject omission (just one type of omission) in the dataset was 2.42\% in the English subset and 16.51\% in Chinese.
It indicates the wide existence of omission in spoken dialogues.

\subsubsection{Coreference}
Pronouns can be used to refer to what is mentioned before in spoken dialogues, which is called coreference.
Two examples are shown in Figure~\ref{fig:example_overview}.
A study~\citep{DBLP:conf/acl/SuSZSHNZ19} shows that coreference occurs in 33.5\% of Chinese daily conversations. 
Statistically,
we use the tools~\citep{spacy_usage,jieba} to count the number of pronouns and find that more than 69.60\% English dialogues and 63.67\% Chinese ones have coreference.
Such high usage of pronouns suggests that coreference is frequent in spoken dialogues, either in English or Chinese.

\subsubsection{Multi-turn Interaction}
Commonly, one speaker interacts with the other in multiple turns in conversation~\citep{DBLP:conf/kdd/LinWHSSL22}.
Statistically, in the Chinese dataset collected from human conversations, speakers switch an average of 270 times per dialogue. In the English dataset, the average number of speaker turns per dialogue is 331.
Furthermore, the MagicData-RAMC~\citep{DBLP:conf/interspeech/YangCLYYCXJZZX022} dataset, also collected from human conversations, has an average of 135 turns per dialogue.
It indicates that multi-turn interactions are important in spoken conversation.

\subsection{Benchmark Dataset Design}\label{sec:BenchmarkDesign}

\subsubsection{Pipeline}

Firstly, we collect real-world spoken dialogues with each phenomenon mentioned in Section~\ref{sec:motivation}.
To cover as many complex conversations as possible, we determine the standard for collection according to the relevant literature (details can be found in Appendix~\ref{appendix:dataset_design}).
With the standard, we collect and extract speech data from web sources and some datasets~\citep{DBLP:conf/emnlp/QuanZCLX20, yu2017formal, shepherd2011want, DBLP:journals/corr/abs-2004-13831, DBLP:journals/tacl/ZhuHZZH20, DBLP:conf/ijcnlp/LiSSLCN17}.

After that, we transfer each real-world spoken dialogue to a unified question instance for the evaluation.
We incorporate each dialogue with a prompt for the evaluation. Different instructions are designed for different phenomena. More details can be found in Section~\ref{sec:prompt_deign}.

\begin{figure*}
    \centering
    \includegraphics[width=1.0\textwidth]{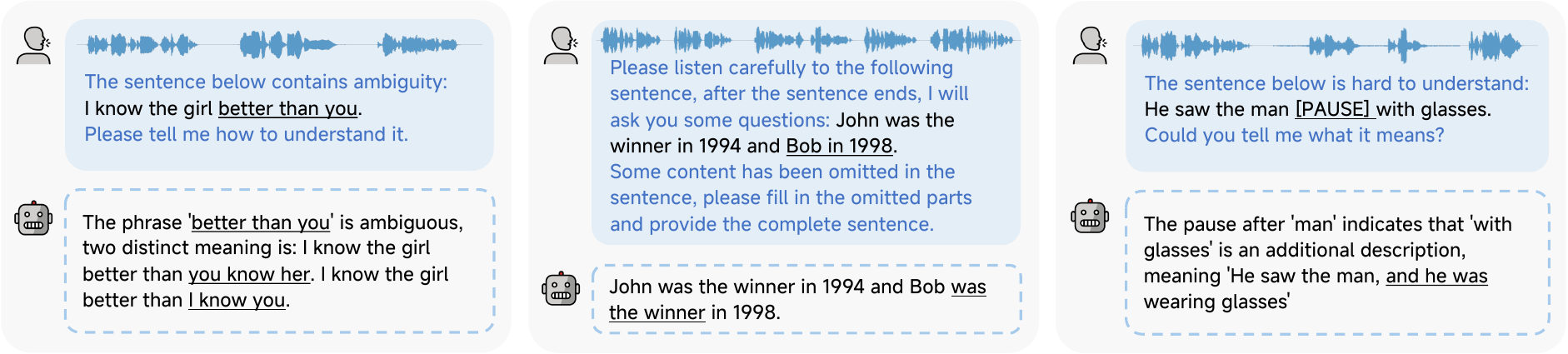}
    \caption{The structure of the data instance. The blue box contains input data in text and audio format, where blue text is the prompt and black text is the dialogue content being questioned. The dashed box contains the reference output, with the underlined portion highlighting the key element. ``[PAUSE]'' represents the pause in the audio.}
    
    \label{fig:am_prompt}
\end{figure*}

For example, the incorporated data instance is shown in Figure~\ref{fig:am_prompt}.
To avoid the influence of irrelevant factors such as timbre and background music, we re-generate each speech data with the tool~\citep{DBLP:journals/corr/abs-2406-02430}, which makes the dialogue content have a unified timbre and no background noise. 

To ensure the quality of the generated speech, we manually check each speech and replace incorrect instances with human voices.
The reference answer in each instance is also manually produced.

\begin{table}[htbp]
    \small
    \centering
    \caption{The number for each category of \cdata. ``zh'' indicates Chinese, and ``en'' indicates English.}
    \label{tab:stat}
    \begin{tabular}{llcc}
        \toprule
        \textbf{Category} & \textbf{Subcategory} & \textbf{zh} & \textbf{en} \\
        \midrule
        \multirow{2}{*}{\textbf{\cambigdata}} 
          & Phonological    & 37  & 29 \\
          & Semantic        & 118 & 51 \\
        \midrule
        \multirow{3}{*}{\textbf{\ccontextdata}}
          & Omission               & 70  & 102 \\
          & Coreference            & 60  & 540 \\
          & Multi-turn Interaction & 38  & 34 \\
        \bottomrule
    \end{tabular}
    
\end{table}

We divide the \cdata into \cambigdata (phonological and semantic ambiguity) to evaluate the ability on ambiguity and \ccontextdata (omission, coreference, and multi-turn interaction) to evaluate the ability on context-dependency (thus the ambiguous dialogues are removed in \ccontextdata). The number of each category is presented in Table~\ref{tab:stat}. There are 1,079 instances in the \ccc, comprising 1,586 audio-text paired samples. The number of audio-text pairs exceeds the number of instances because multi-turn dialogues contain multiple samples.

\subsubsection{Data Instance Construction}\label{sec:prompt_deign}

To evaluate SDM's performance across different complex phenomena, we design specialized instructions for each category. The complete set of instructions and annotation details are provided in Appendix~\ref{sec:detailed_exemplar}.

\noindent\textbf{Phonological Ambiguity}:
The phonological ambiguity evaluates both the comprehension and generation capabilities of the SDM.
For comprehension assessment, we instruct the SDM that the input contains potentially ambiguous phonological features and request a detailed interpretation.
For generation assessment, we explicitly indicate the presence of incorrect phonological features (e.g., pauses, intonation) and prompt the SDM to generate a corrected response with appropriate prosodic markers.

\noindent\textbf{Semantic Ambiguity}:
We inform the SDM that the meaning of the instance is unclear and instruct the SDM to provide a detailed explanation.

\noindent\textbf{Omission}:
Our assessment focuses on two capabilities,
(1) Detection: Instruct the SDM to identify if there are missing elements in the dialogues.
(2) Completion: Inform that some content is omitted and instruct SDM to provide the completed sentence with the omission.

\noindent\textbf{Coreference}:
We evaluate two related skills,
(1) Detection: Instruct the SDM to identify if there is any coreference in the instance.
(2) Resolution: Inform that the coreference phenomenon exists in the dialogue and instruct SDM to provide the coreference relationship.

\noindent\textbf{Multi-turn Interaction}:

After the real-world multi-turn dialogues, we 
repeat the initial question and instruct SDM to provide the identical answer as the previous one.

\section{Experiment Settings and Evaluation}

\subsection{Experimental Settings}

We select end-to-end SDMs instead of cascaded ones because the latter are unable to retain the phonological features such as press, pause, and intonation during ASR.

For the SDMs (i.e., Freeze-Omni, LLaMA-Omni, VITA-Audio, and MooER-Omni) that do not natively support multi-turn interaction,
we concatenate the dialogue history in sequence before the current input 
in the evaluation.
The real-time full-duplex model (i.e., Moshi) interrupts the input audio when provided with dialogue history, resulting in responses beyond the posed questions. As it cannot be evaluated in the same setting of multi-turn interaction as others, it is not fair to be compared and thus not chosen.
Note that some models (i.e, LLaMA-Omni and Moshi) do not support Chinese; therefore, they are evaluated only in English.

\begin{table*}[htbp]
\centering
\tiny
\setlength{\tabcolsep}{2pt}
\caption{Accuracy (\%) of different SDMs on the Chinese (``zh'') or English (``en'') dialogue data subset of \ccc.}
\label{tab:combined_results}
\begin{tabular}{l *{11}{cc}}
\toprule
\multirow{2}{*}{\textbf{Category}}
& \multicolumn{2}{c}{\textbf{Freeze-Omni}}
& \multicolumn{2}{c}{\textbf{GLM-4-Voice}}
& \multicolumn{2}{c}{\textbf{\makecell{GPT-4o-\\Audio-Prev.}}}
& \multicolumn{2}{c}{\textbf{Kimi-Audio}}
& \multicolumn{2}{c}{\textbf{\makecell{LLaMA-\\Omni}}}
& \multicolumn{2}{c}{\textbf{\makecell{MooER-\\Omni}}}
& \multicolumn{2}{c}{\textbf{Moshi}}
& \multicolumn{2}{c}{\textbf{\makecell{Qwen2.5-\\Omni}}}
& \multicolumn{2}{c}{\textbf{Step-Audio}}
& \multicolumn{2}{c}{\textbf{VITA-Audio}}
& \multicolumn{2}{c}{\textbf{Overall}} \\
\cmidrule(lr){2-3} \cmidrule(lr){4-5} \cmidrule(lr){6-7} \cmidrule(lr){8-9} \cmidrule(lr){10-11} \cmidrule(lr){12-13} \cmidrule(lr){14-15} \cmidrule(lr){16-17} \cmidrule(lr){18-19} \cmidrule(lr){20-21} \cmidrule(lr){22-23} 
& \textbf{zh} & \textbf{en} & \textbf{zh} & \textbf{en} & \textbf{zh} & \textbf{en} & \textbf{zh} & \textbf{en} & \multicolumn{2}{c}{\textbf{en}} & \textbf{zh} & \textbf{en} & \multicolumn{2}{c}{\textbf{en}} & \textbf{zh} & \textbf{en} & \textbf{zh} & \textbf{en} & \textbf{zh} & \textbf{en} & \textbf{zh} & \textbf{en} \\
\midrule
Phonological & 16.22 & 8.62 & 18.92 & 27.59 & \textbf{29.73} & \textbf{53.45} & 20.27 & 46.55 & \multicolumn{2}{c}{15.52} & 20.27 & 18.97 & \multicolumn{2}{c}{10.34} & 27.03 & 48.28 & 22.97 & 29.31 & 8.11 & 31.03 & 20.44 & 28.97 \\
Semantic & 1.69 & 11.76 & 2.54 & 15.69 & 5.93 & \textbf{70.59} & 4.24 & 29.41 & \multicolumn{2}{c}{12.75} & 2.12 & 46.08 & \multicolumn{2}{c}{9.80} & \textbf{6.78} & 32.35 & 5.08 & 21.57 & 3.39 & 18.63 & 3.97 & 26.86 \\
\addlinespace
\textbf{\cambigdata} & 8.96 & 10.19 & 10.73 & 21.64 & \textbf{17.83} & \textbf{62.02} & 12.25 & 37.98 & \multicolumn{2}{c}{14.13} & 11.19 & 32.52 & \multicolumn{2}{c}{10.07} & 16.90 & 40.31 & 14.03 & 25.44 & 5.75 & 24.83 & 12.21 & 27.91 \\
\midrule
Omission & 4.29 & 6.86 & 5.71 & 6.37 & \textbf{44.29} & \textbf{16.18} & 29.29 & 10.29 & \multicolumn{2}{c}{5.88} & 32.14 & 4.90 & \multicolumn{2}{c}{2.94} & 27.86 & 15.20 & 17.86 & 10.78 & 6.43 & 7.84 & 20.98 & 8.73 \\
Coreference & 10.83 & 47.22 & 16.67 & 68.98 & 54.17 & \textbf{91.11} & 40.00 & 87.41 & \multicolumn{2}{c}{56.94} & 32.50 & 36.02 & \multicolumn{2}{c}{24.63} & \textbf{55.83} & 68.15 & 50.83 & 57.31 & 33.33 & 74.81 & 36.77 & 61.26 \\
Multi-turn & 11.84 & 44.12 & 10.53 & 58.82 & 13.16 & 47.06 & / & / & \multicolumn{2}{c}{55.88} & 63.16 & 41.18 & \multicolumn{2}{c}{/} & \textbf{82.89} & \textbf{95.59} & 7.89 & 41.18 & 63.16 & 60.29 & 36.09 & 55.51 \\
\addlinespace
\textbf{\ccontextdata} & 8.99 & 32.73 & 10.97 & 44.73 & 37.20 & 51.45 & 34.64 & 48.85 & \multicolumn{2}{c}{39.57} & 42.60 & 27.37 & \multicolumn{2}{c}{13.79} & \textbf{55.53} & \textbf{59.64} & 25.53 & 36.43 & 34.31 & 47.65 & 31.22 & 40.22 \\
\midrule
\textbf{Overall} & 8.97 & 23.72 & 10.87 & 35.49 & 29.45 & \textbf{55.68} & 23.45 & 43.42 & \multicolumn{2}{c}{29.39} & 30.04 & 29.43 & \multicolumn{2}{c}{11.93} & \textbf{40.08} & 51.91 & 20.93 & 32.03 & 22.88 & 38.52 & 23.33 & 35.15 \\
\bottomrule
\end{tabular}

\end{table*}

\subsection{LLM-based Evaluation}

\paragraph{Preprocessing}

Most SDMs output both audio and corresponding text simultaneously. For the model (i.e., Moshi) without generating corresponding text, we convert the audio to text using Whisper~\citep{DBLP:conf/icml/RadfordKXBMS23}.

\paragraph{Evaluation Method}

We adopt different methods for different categories in the dataset, \cdata.
For most tasks, except for generating audio with correct phonological features in the phonological ambiguity phenomenon, we evaluate the transcribed text from the audio. 
This is because phonological features in the response do not affect the comparison results with the reference, so evaluating the text alone is sufficient. 
For the task of generating audio with correct phonological features, we evaluate the audio output manually, as it requires examining phonological features that cannot be captured by the transcribed text.

For the evaluation based on transcribed text, we design an automatic LLM-based evaluation method following the paradigm of LLM-as-a-judge~\citep{DBLP:journals/corr/abs-2411-15594}. \gptfouro~\citep{GPT-4o} and \dpskrone~\citep{DBLP:journals/corr/abs-2501-12948} are selected as LLM judge due to their great performance in reasoning~\citep{DBLP:journals/corr/abs-2501-12948}. LLM judges are used to compare the SDM output with the reference and determine the correctness.
Moreover, we divide the evaluation task into smaller steps, instructing LLM judges with the prompts that are listed in the repository\footnote{\url{https://step-out.github.io/C3-web}}.
For the evaluation based on the audio, three human experts are required to label whether each of the SDM outputs is correct, and we use a voting strategy to make the final decision for each generated response.

The accuracy (i.e., the proportion of instances judged correct out of the total number of instances) is regarded as the metric. 

\paragraph{Reliability Analysis}

To validate the reliability of our designed automatic evaluation method, we first conduct a human evaluation on the generated responses by \gptfouroa for \cdata.
Following best practice for the human evaluation~\citep{LeeGMWK19}, three human experts manually label whether each response is correct. If the labels from all experts are not the same, the majority label is chosen as the reference result. 

After the human evaluation, we computed the Pearson~\citep{cohen2009pearson}, Spearman~\citep{DBLP:journals/concurrency/XiaoYER16}, and Kendall~\citep{abdi2007kendall} correlation coefficients to quantify the consistency between LLM judges and humans. 
All the coefficients' values are more than 0.87 in either the English or Chinese subset, either for \dpskrone or \gptfouro as LLM judge (detailed numbers can be found in Appendix~\ref{sec:correlation}). It demonstrates that LLM judges have high consistency with humans in each subset for the two LLMs.
Moreover, all p-values of the correlation coefficients are less than 0.001, which means the consistency is significant. These statistical results validate the reliability of our automatic evaluation method. 

\section{Experimental Results and Findings}

\subsection{Experimental Results}

To mitigate bias between \dpskrone and \gptfouro, we compute the average of their accuracies as the final result, as shown in Table~\ref{tab:combined_results}. 
The SDMs perform differently across different languages and phenomena.

As shown in Table~\ref{tab:combined_results}, the gap between English and Chinese exceeds 8\% across each phenomenon, indicating that SDMs exhibit varying capabilities depending on the language. 
Meanwhile, in the English subset, \gptfouroa significantly outperforms other models, achieving an overall accuracy of 55.68\%, while the average performance of all SDMs is only 35.15\%. In contrast, in the Chinese subset, Qwen2.5-Omni stands out as the top-performing SDM, achieving an overall accuracy of 40.08\%, while the average performance of all SDMs is 23.33\%.
The gap between Chinese and English in top performances and overall scores further highlights the differing strengths of SDMs across languages.

Within the same language, the performance gap between the strongest and weakest phenomena is over 9 times (for Chinese) and 6 times (for English), suggesting that SDMs vary in their strengths across different phenomena.

\begin{figure}[t]
    \centering
    
    \includegraphics[width=0.8\linewidth]{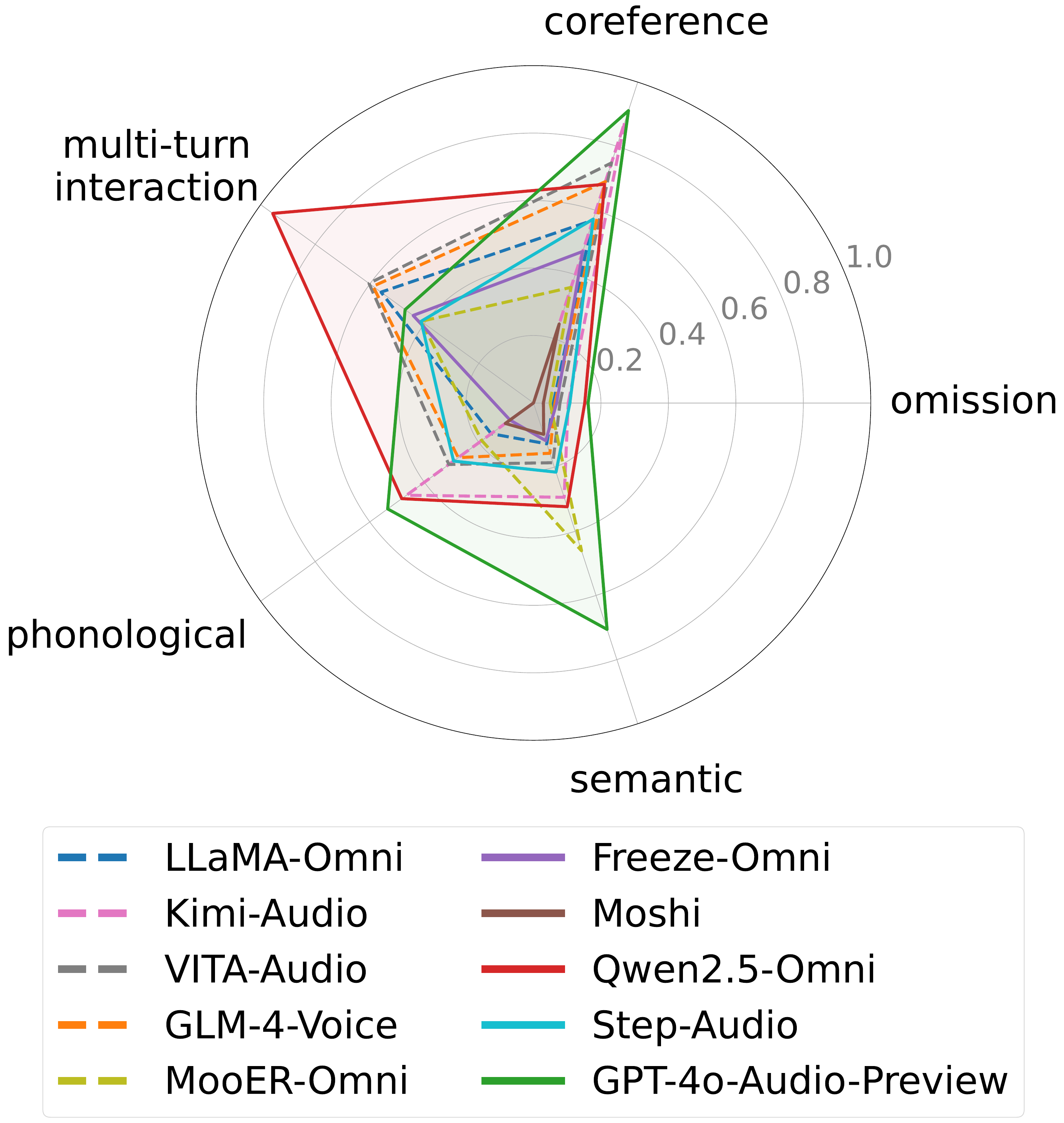}
    \caption{Radar charts depicting the accuracies of each SDM on the English subset of \cdata.}
    \label{en_gpt}
    
\end{figure}

\begin{figure}[t]
    \centering
    \includegraphics[width=0.8\linewidth]{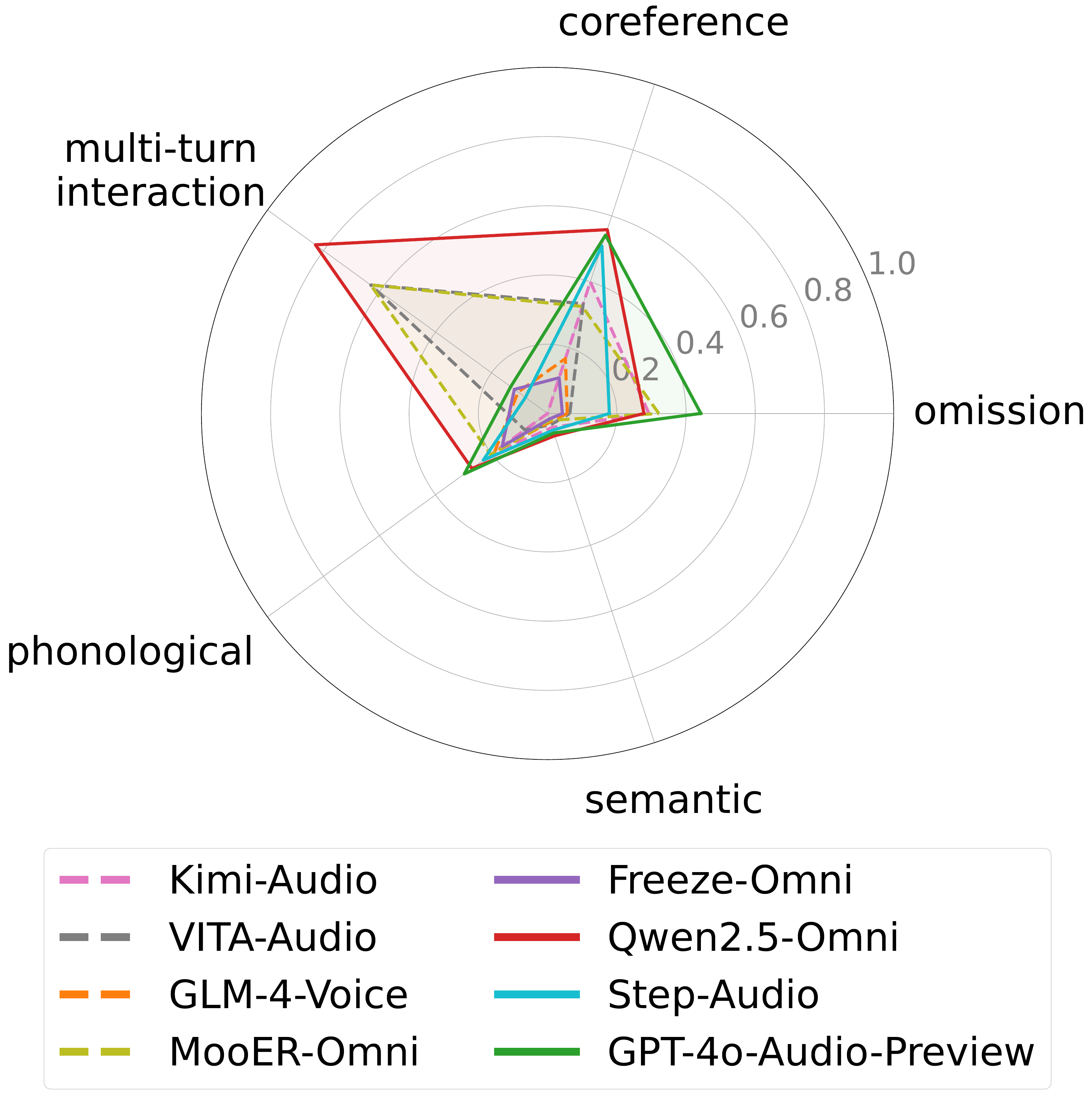}
    \caption{Radar charts depicting the accuracies of each SDM on the Chinese subset of \cdata.}
    \label{cn_gpt}
\end{figure}

To illustrate the performance in handling different phenomena, radar charts are presented in Figure~\ref{en_gpt} and Figure~\ref{cn_gpt}. 
As shown in Figure~\ref{en_gpt}, \gptfouroa has the largest green area compared to the others, which validates its top performance. 
In the dimension of multi-turn interaction, \gptfouroa scores significantly lower than Qwen2.5-Omni, indicating a weakness of the model. Although the overall scores of the top two SDMs, \gptfouroa at 55.68\% and Qwen2.5-Omni at 51.91\%, are relatively close, each model exhibits distinct advantages. 
As shown in Figure~\ref{cn_gpt}, Qwen2.5-Omni excels in multi-turn interaction, with a sharp accuracy gap over other SDMs. The performance of the SDMs further highlights their varying strengths across different phenomena.
Note that the detailed results from each LLM judge can be found in Appendix~\ref{sec:detailed_result}.

To further investigate the ability to handle dialogues with omission and coreference,
two tasks, including detection and completion (resolution), are provided for the evaluation.
The final results of these two tasks are presented in Table~\ref{tab:merged_analysis}.

\begin{table*}[htbp]
    \scriptsize
    \centering
    \setlength{\tabcolsep}{3pt}
    \caption{Accuracy (\%) of omission and coreference phenomena.}
    \label{tab:merged_analysis}
    \begin{tabular}{cccccccccccccc}
    \toprule
    \multirow{1}{*}{\textbf{Phenomenon}} & \multirow{1}{*}{\textbf{Ability}} & \multirow{1}{*}{\textbf{Lang}}&
    \textbf{\makecell{Freeze-\\Omni}}& \textbf{\makecell{GLM-4-\\Voice}}& \textbf{\makecell{GPT-4o-\\Audio-Prev.}}& \textbf{\makecell{Kimi-\\Audio}}& \textbf{\makecell{LLaMA-\\Omni}}& \textbf{\makecell{MooER-\\Omni}}& \textbf{Moshi}& \textbf{\makecell{Qwen2.5-\\Omni}}& \textbf{\makecell{Step-\\Audio}}& \textbf{\makecell{VITA-\\Audio}} & \textbf{Overall} \\
    \midrule

    \multirow{4}{*}{Omission}
    & \multirow{2}{*}{Detection} & zh & 8.57 & 10.00 & \textbf{82.86} & 52.86 & / & 61.43 & / & 48.57 & 32.86 & 8.57 & 38.65 \\
    & & en & 8.82 & 4.90 & \textbf{13.73} & 12.75 & 6.86 & 6.86 & 3.92 & 11.76 & 7.84 & 6.86 & 8.57 \\
    \cmidrule(lr){2-14}
    & \multirow{2}{*}{Completion} & zh & 0.00 & 1.43 & 5.71 & 5.71 & / & 2.86 & / & \textbf{7.14} & 2.86 & 4.29 & 3.75 \\
    & & en & 4.90 & 7.84 & \textbf{18.63} & 7.84 & 4.90 & 2.94 & 1.96 & \textbf{18.63} & 13.73 & 8.82 & 8.98 \\
    \midrule

    \multirow{4}{*}{Coreference}
    & \multirow{2}{*}{Detection} & zh & 20.00 & 33.33 & 63.33 & 60.00 & / & 58.33 & / & \textbf{86.67} & 70.00 & 56.67 & 56.41 \\
    & & en & 57.59 & 83.89 & 95.37 & \textbf{97.04} & 78.52 & 25.93 & 35.37 & 70.56 & 63.33 & 87.59 & 69.61 \\
    \cmidrule(lr){2-14}
    & \multirow{2}{*}{Resolution} & zh & 1.67 & 0.00 & \textbf{45.00} & 20.00 & / & 6.67 & / & 25.00 & 31.67 & 10.00 & 17.00 \\
    & & en & 36.85 & 54.07 & \textbf{86.85} & 77.78 & 35.37 & 46.11 & 13.89 & 65.74 & 51.30 & 62.04 & 52.17 \\
    \bottomrule
    \end{tabular}
\end{table*}

\subsection{Experimental Findings}

\subsubsection{Ambiguity Is Difficult for SDMs Especially Semantic Ones in Chinese}

As shown in Table~\ref{tab:combined_results}, SDMs achieve overall accuracies of 12.21\% (Chinese) and 27.91\% (English) on \cambigdata, significantly lower than the 31.22\% (Chinese) and 40.22\% (English) observed on \ccontextdata. The performance gap of over 10 percentage points in both languages suggests that ambiguity presents greater challenges for SDMs.
Specifically, the overall accuracy in semantic ambiguity is only 3.97\% in Chinese, compared to 26.86\% in English. This pronounced disparity (exceeding a six-fold difference) underscores the challenges of processing semantic ambiguity in Chinese.

Additionally, the difference in accuracies for phonological ambiguity, 20.44\% (Chinese) and 28.97\% (English), exceeds an 8\% gap. 
The exception is MooER-Omni, which has a gap of less than 1.5 percentage points. This contrast highlights MooER-Omni's cross-linguistic ability to handle phonological ambiguity.

\subsubsection{Processing Omission Is the Most Difficult in Context-Dependency}

Table~\ref{tab:combined_results} shows that, except for \gptfouroa and Step-Audio in Chinese, all SDMs have the smallest accuracy when dealing with the omission phenomenon among \ccontextdata. 
This indicates that omission is the most difficult phenomenon for SDMs to handle in context-dependent dialogues.

Dealing with spoken dialogues with omission or coreference requires both detection and completion (or resolution). 
To investigate the abilities of SDMs at a granular level, we compare the accuracies of each ability as shown in Table~\ref{tab:merged_analysis}. 
In omission, most SDMs have higher accuracy in detection than in completion.
This suggests that although the omission is pointed out, the SDMs could not fully understand and thus complete the missing part.
The exception is GLM-4-Voice, \gptfouroa, Qwen2.5-Omni, Step-Audio, and VITA-Audio in English. With the prompt of the omission phenomenon, these five SDMs can complete more than what they can detect on their own.
In coreference, the finding is similar. Most SDMs have higher accuracy in detection than resolution, indicating that although the coreference is pointed out, the SDMs cannot fully understand and resolve it.
The exception is MooER-Omni in English, which performs better when pointing out coreference.
The above findings teach us that pointing out the phenomenon in dialogue can be helpful for some SDMs, but most of them benefit only slightly.

We also find that most SDMs demonstrated higher accuracy in dealing with coreference resolution than omission completion. 
The different performances of these two phenomena can be inferred:
In the coreference phenomenon, both the pronoun and the antecedent are present in the sentence. The SDM can replace the pronoun with the antecedent by understanding the sentence. However, in the omission phenomenon, the omitted content is not present in the sentence. To complete the omitted parts, the SDM should not only understand each component's meaning but also generate non-existent components. Therefore, resolving the omission phenomenon is more difficult for SDMs than resolving the coreference phenomenon.

Moreover, we observe that most SDMs exhibit low accuracies (below 65\%) in multi-turn interactions, whereas Qwen2.5-Omni achieves significantly higher accuracy, with 82.89\% for Chinese and 95.59\% for English, outperforming the other models.

\subsubsection{Complex Dialogues in Chinese Are More Difficult than Ones in English}

As shown in Table~\ref{tab:combined_results}, 
the overall accuracies for both \cambigdata and \ccontextdata are higher in English (27.91\% and 40.22\%) than in Chinese (12.21\% and 31.22\%). The difference exceeds nine percentage points, indicating that, generally, SDMs perform better in English dialogues.

Specifically, in each phenomenon, the overall accuracy in English is higher, except for omission, suggesting that English phenomena are generally easier for SDMs than their Chinese counterparts.

Specifically, as shown in Table~\ref{tab:combined_results}, most SDMs demonstrate higher accuracy in English than in Chinese. For instance, Freeze-Omni and GLM-4-Voice achieve accuracies of 23.72\% and 35.49\% in English, more than double their performance in Chinese (8.97\% and 10.87\%). This substantial gap highlights the need for enhanced cross-linguistic capabilities in current SDMs.

\noindent\textbf{Summary:}
These findings suggest that the choice of SDM should depend on the specific situation, such as the phenomenon or language.

\section{Conclusion}

In this work, we introduce a new benchmark, \ccc, to evaluate SDMs' capabilities in handling various complex conversations. 
Our empirical study reveals five important phenomena in spoken dialogues that are not fully explored in previous works.
With our designed dataset, \cdata, and LLM-based evaluation method, SDMs can be evaluated more comprehensively.
Furthermore, we conduct experiments on ten SDMs. The results point out different difficulties in processing these complex phenomena in different languages. 

We believe that \ccc, including real and complex challenges in spoken dialogues, is helpful for researchers to achieve natural and intelligent spoken interaction with humans.
In the future, we will collect more language dialogues into \cdata. 

\section*{Limitations}

There are two limitations to this work:
First, the five complex phenomena discussed in this paper are not limited to English and Chinese; they have significant potential for other languages.
Second, there is potential bias among human experts who evaluate the outputs of SDMs. To mitigate this bias, we employ a voting mechanism.
\bibliography{reference}

\appendix

\section{Appendix}

\subsection{Deatils of Dataset Design}~\label{appendix:dataset_design}
Based on our empirical study, we optimize the data construction process and introduce criteria to ensure the quality of the dataset (Section~\ref{sec:BenchmarkDesign}). The specific filtering criteria are as follows:

\noindent\textbf{Phonological Ambiguity}: The intended meaning of the instances is ambiguous, caused by phonological features including heteronym, heterograph, stress, intonation, pause, and tone-only difference.

\noindent\textbf{Semantic Ambiguity}: The sentence contains lexical or syntactic ambiguities. More specifically, lexical ambiguity means it contains polysemous words, while syntactic ambiguity means the phrase or sentence can be parsed in more than one way grammatically.

\noindent\textbf{Omission}: The instance omits part of the utterance, and the omission must be inferred from the surrounding context or common knowledge. More specifically, the omission can be the word: subject, verb, or object, commonly understood by speakers.

\noindent\textbf{Coreference}: The instance uses pronouns (e.g., he, she, that) or phrases (e.g., the former, the boy) to refer to specific entities mentioned in the dialogue.

\noindent\textbf{Multi-turn Interaction}: The instance includes at least five turns with speaker alternation, and semantic dependencies are across dialogue turns.

The dataset design process for each phenomenon is described in detail below.

\subsubsection{Phonological Ambiguities}

\paragraph{Understanding}
In the Chinese subset, ambiguities arise from four types of characteristics: pause, heteronym, heterograph, and syllable with different tones. In the English subset, ambiguities result from four types of characteristics: heterograph, pause, stress, and intonation. During the manual review of the TTS-generated audio, we find that the audio quality in the Chinese dataset is poor for the pause and heteronym characteristics, and in the English dataset for the pause and stress characteristics. Consequently, these four parts of the data are re-recorded manually. The final bilingual dataset contains ambiguous questions with audio and text modalities, and the corresponding textual reference answers. The form of the other dataset is also the same.

\paragraph{Generating}
In addition, we develop data that tests SDM's ability to generate dialogues with phonetic characteristics. This data is derived from the understanding phonological ambiguities dataset. 
An exception is heterographs, which are not included in the generation evaluation, as the reference audio for homophones remains the same.
The ambiguous sentences remain unchanged, but the prompts are changed in different characteristics and different languages. For pauses, stresses, and intonations, the evaluation involves inputting the meaning of the ambiguous sentence and assessing whether the SDM produces these phonological features appropriately. For heteronyms and syllables with different tones, the evaluation involves inputting sentences with incorrect pronunciations and assessing whether the SDM can correct them based on context.

\subsubsection{Semantic Ambiguities}

This subset of \cambigdata examines the ability of SDM to process semantic ambiguities. We first identify the types of semantic ambiguities to collect data from relevant literature~\citep{rodd2018lexical, taha1983types, dai2021syntactic, lasheiky2024semantic}. Subsequently, we manually gather data from various websites, including ambiguous sentences and their interpretations. The data are then organized using a standardized prompt, instructing the SDM to provide interpretations of the ambiguous sentences. Finally, the data are converted from text to audio and checked manually for quality.

The Chinese dataset encompasses ambiguities arising from unclear pronominal reference, polysemy, unclear modification scope, unclear part of speech, and unclear subject-object relationship. To ensure optimal audio quality, the first two instances are enhanced with human-voiced recordings, and the remaining are generated by TTS. The English dataset includes lexical ambiguities stemming from unclear parts of speech and polysemy, as well as syntactic ambiguities resulting from unclear pronominal reference and unclear modification scope.

\subsubsection{Omission}

This section of the dataset examines SDM's ability to understand comprehension difficulties in dialogues caused by the omission phenomenon. The Chinese dataset is based on the RISAWOZ dataset~\citep{DBLP:conf/emnlp/QuanZCLX20}, a text dataset specifically designed to study coreference and omission phenomena. The selected portion of the RISAWOZ dataset contains multi-turn dialogues with 1, 3, or 5 sentences, and provides annotations for omission and coreference in each dialogue. We retain the segments of each multi-turn dialogue from the beginning up to the point where omission occurs and add prompts to query SDM to construct the dataset. The English portion is manually extracted from relevant literature~\citep{yu2017formal, shepherd2011want}, and data containing the omission phenomenon is constructed with corresponding prompts to query SDM. Unlike the Chinese dataset, which is in the form of multi-turn dialogues, the English dataset is in the form of a single sentence.

For both the Chinese and English datasets, reference answers that supplement the omitted content are provided to enable comparison with SDM's responses. We construct two questions in the prompt for the data: The first question asks the SDM to determine whether there is an omission phenomenon in the input audio, and the second question informs the SDM of the existence of an omission phenomenon in the input and requests the SDM to complete the omitted content. The two questions are independent of each other. To prevent overlap with the ambiguous contexts dataset, we exclude the omission phenomenon that would cause ambiguity during data selection.

\subsubsection{Coreference}

This section of the dataset assesses SDM's ability to comprehend difficulties in dialogues arising from the coreference phenomenon. The Chinese dataset is based on the RISAWOZ dataset~\citep{DBLP:conf/emnlp/QuanZCLX20} and employs a similar methodology to the omission section, resulting in multi-turn dialogue data instances, each comprising 1, 3, or 5 sentences. The dataset includes reference answers that resolve coreference by replacing pronouns with their referents, thereby eliminating the coreference phenomenon, to serve as a standard for comparing SDM's responses.

The English dataset is constructed based on the Winograd Schema Challenge dataset~\citep{DBLP:journals/corr/abs-2004-13831}. Each data instance comprises a sentence and a multiple-choice question targeting the referent of a pronoun, with two potential answers provided. The dataset also includes the correct answer to each coreference question. The referents of these pronouns are easily confused, necessitating a deep understanding of the sentence's meaning as well as robust commonsense knowledge and reasoning capacity to determine the correct answer. To ensure the dataset's quality and clarity regarding the pronouns in question, we filter out instances where the pronoun appears more than once in the sentence.

Consistent with the omission dataset, to avoid overlap with ambiguous contexts, we exclude coreference phenomena that could introduce ambiguity. We then task the SDM with addressing two distinct queries: first, to verify the presence of the coreference phenomenon, and second, to deliver the outcomes following coreference resolution.

\subsubsection{Multi-turn Interaction}

To evaluate the model's ability to track conversation history, we ensured that the assessment of SDM is conducted in a multi-turn conversational format. The criteria for collecting data are that the dialogues must be multi-turn. The Chinese dataset is based on the CrossWoz dataset~\citep{DBLP:journals/tacl/ZhuHZZH20}, which covers multiple domains including tourist attractions, hotels, restaurants, subways, and taxis. The English dataset is derived from the DailyDialog dataset~\citep{DBLP:conf/ijcnlp/LiSSLCN17}, which is artificially constructed with minimal noise and encompasses a variety of everyday conversational scenarios. Since our method of evaluating SDM involves posing the first question in the dialogue, we ensured that the first sentence of each dialogue is a question when filtering out the dataset. Defining a single input to the SDM and its corresponding response as one turn of dialogue, the Chinese dataset features dialogues with a maximum of 16 turns and an average of 9.68 turns, whereas the English dataset has a maximum of 9 turns and an average of 6.21 turns.

In the dataset, only the content input by the user to SDM is provided, while the responses of SDM are generated by the SDM being evaluated. This subset of \ccontextdata examines SDM's ability to remember the content of multi-turn dialogues and to utilize the conversation history to generate current responses when processing dialogues. Therefore, after the dialogue concludes, we revisit the first question in the dialogue and request that SDM respond to that question again. If the final response provided by SDM is consistent with the initial response and the intervening question-and-answer content, it is considered to have good capability to process multi-turn interaction. During the evaluation process, if the SDM being evaluated only provides single-turn dialogue capability, we concatenate the previous question-and-answer pairs to manually construct the conversation history for each input.

\subsection{Detailed Structure and Exemplars of Dataset} 
\label{sec:detailed_exemplar}

The annotation details for each phenomenon are as follows:

\noindent\textbf{Phonological Ambiguity}: Different meanings are annotated for each sentence, along with the correct phonological features, including pronunciation, intonation, stress position, and pause position.

\noindent\textbf{Semantic Ambiguity}: Semantic ambiguity is divided into lexical and syntactic ambiguity. For lexical ambiguity, different meanings of the same word are annotated. For syntactic ambiguity, different interpretations of the same semantic structure are annotated.

\noindent\textbf{Omission}: The omitted parts are annotated based on context and common sense.

\noindent\textbf{Coreference}: The word or phrase referred to by the pronoun is annotated based on context and common sense.

\noindent\textbf{Multi-turn Interaction}: Multi-turn dialogues do not require annotation, as the reference answer is determined by the SDM's output.

\begin{figure*}[p]
  \centering
  \includegraphics[width=0.7\textwidth]{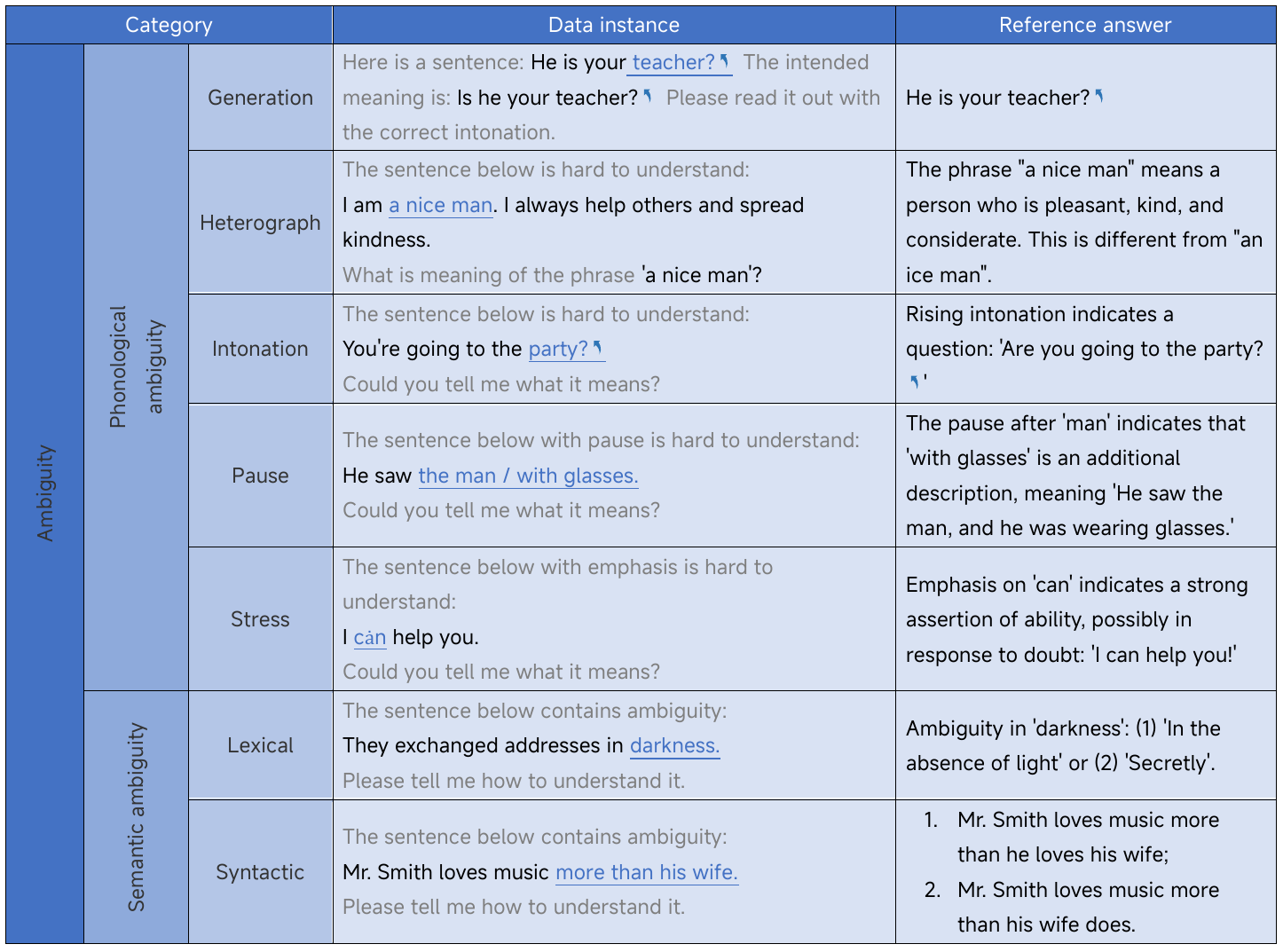}
  \caption{The figure delineates the structure and exemplars of the English Ambiguous subset within the dataset.}
  \label{fig:en-am}
\end{figure*}

\begin{figure*}[p]
  \centering
  \includegraphics[width=0.7\textwidth]{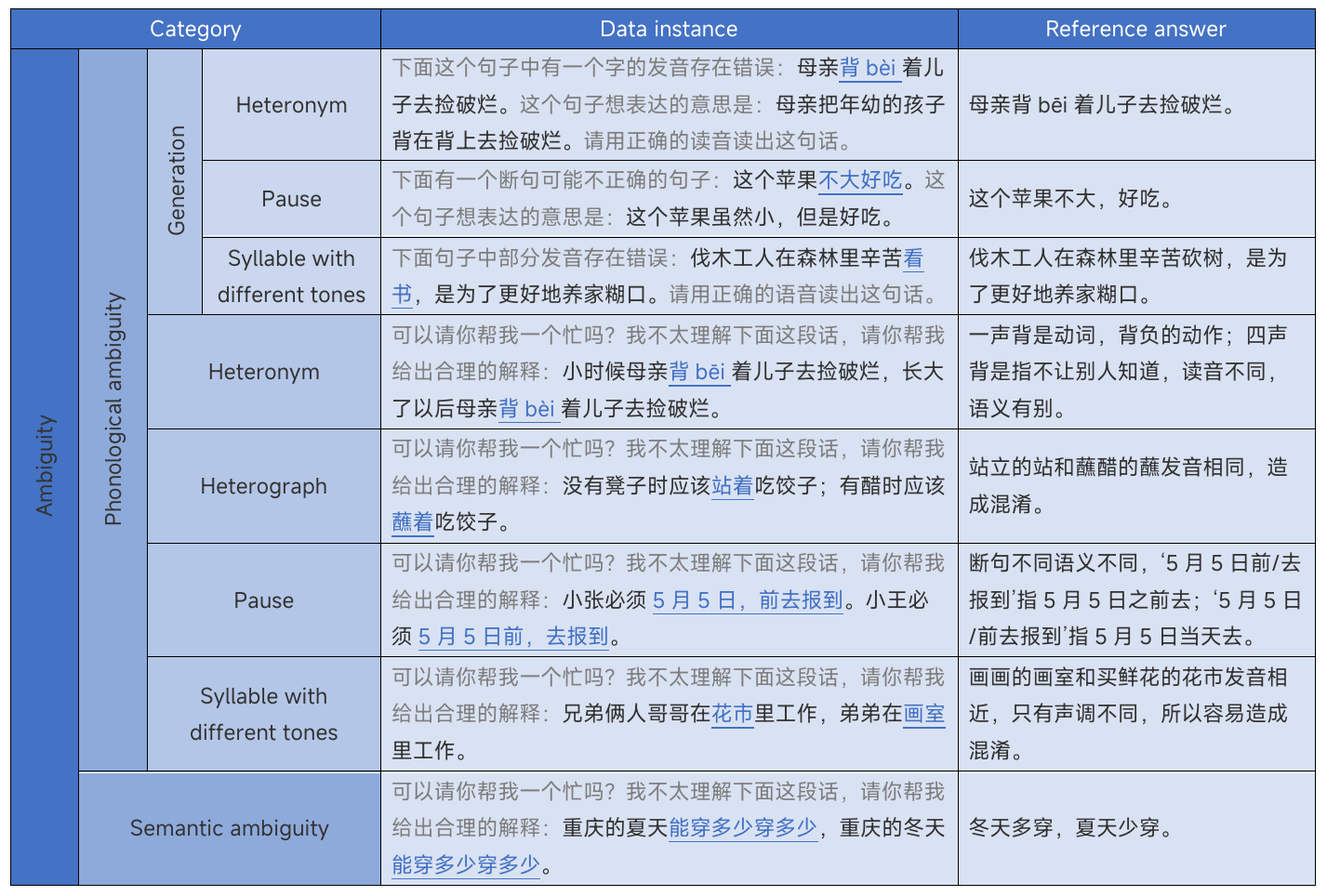}
  \caption{The figure delineates the structure and exemplars of the Chinese Ambiguous subset within the dataset.}
  \label{fig:cn-am}
\end{figure*}

\begin{figure*}[p]
  \centering
  \includegraphics[width=0.7\textwidth]{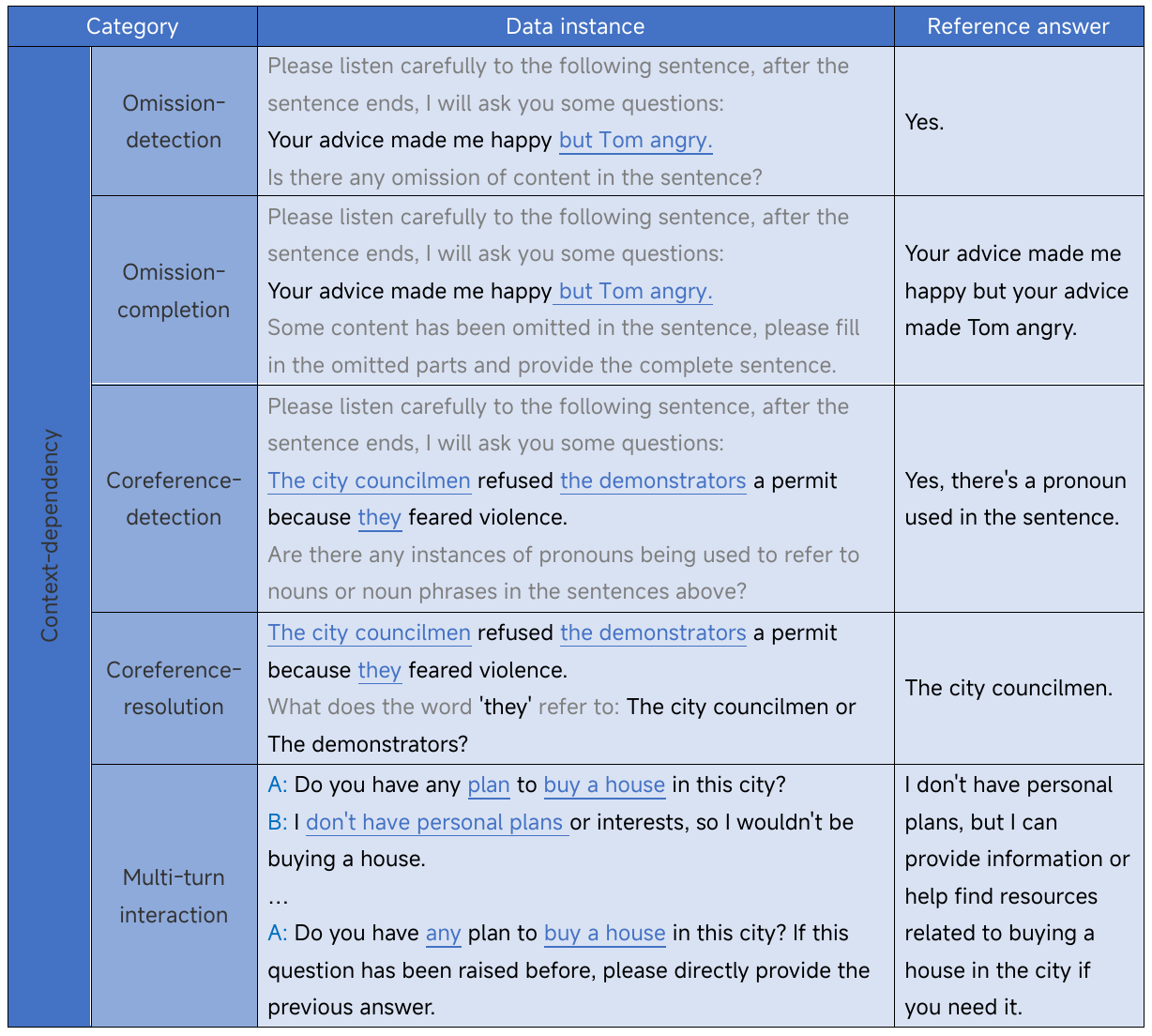}
  \caption{The figure delineates the structure and exemplars of the English Context-Dependency subset within the dataset.}
  \label{fig:en-non}
\end{figure*}

\begin{figure*}[p]
  \centering
  \includegraphics[width=0.7\textwidth]{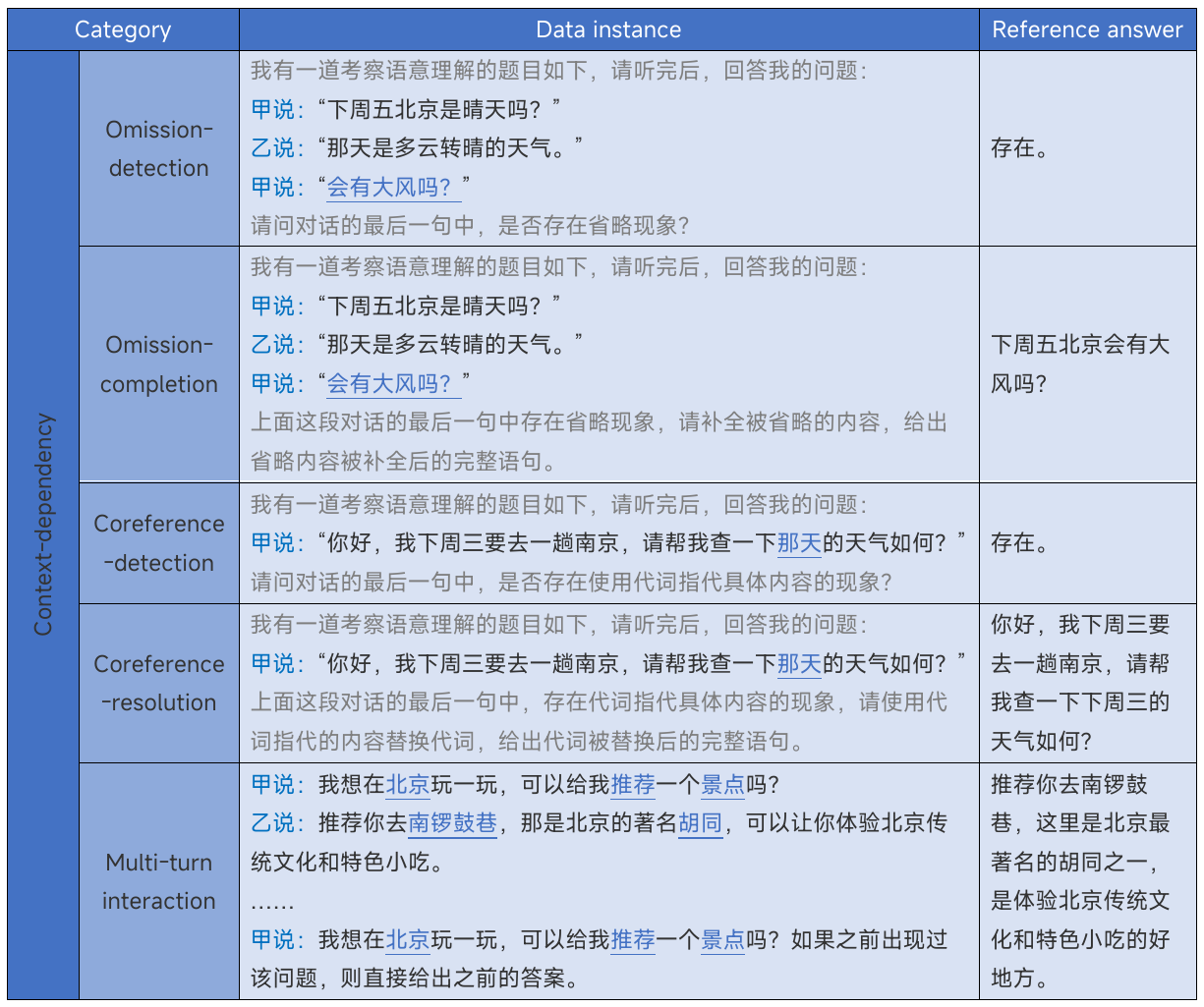}
  \caption{The figure delineates the structure and exemplars of the Chinese Context-Dependency subset within the dataset.}
  \label{fig:cn-non}
\end{figure*}

To provide a more detailed illustration of the contents of each subset within the dataset, 
Figure~\ref{fig:en-am} -~\ref{fig:cn-non}
have been presented. The gray text denotes the invariant segments integral to the dataset's construction, immutable irrespective of variations in the data samples. The underlined blue-highlighted segments indicate the focal areas examined by the SDM, while the non-underlined blue-highlighted portions distinguish the roles of different participants in multi-turn dialogues.

\subsection{Correlation Analysis} 
\label{sec:correlation}
To further illustrate the correlation between LLMs and human evaluations, Table~\ref{tab:correlation} presents three correlation coefficients, while Table~\ref{tab:pvalues} shows their corresponding p - values.

\begin{table*}[htbp]
    \small
    \centering
    \caption{Correlation coefficients between LLM evaluation results and human assessment results}
    \label{tab:correlation}
    \begin{tabular}{lcccc}
        \toprule
        \textbf{Model} & \textbf{Pearson} & \textbf{Spearman} & \textbf{Kendall} & \textbf{Language} \\
        \midrule
        \dpskrone & 0.8969 & 0.8969 & 0.8969 & Chinese \\
        \gptfouro & 0.8886 & 0.8886 & 0.8886 & Chinese \\
        \dpskrone & 0.8739 & 0.8739 & 0.8739 & English \\
        \gptfouro & 0.8940 & 0.8940 & 0.8940 & English \\
        \bottomrule
    \end{tabular}
\end{table*}

\begin{table*}[htbp]
    \small
    \centering
    \caption{p-values for correlation coefficients}
    \label{tab:pvalues}
    \begin{tabular}{lcccc}
        \toprule
        \textbf{Model} & \textbf{Pearson p-value} & \textbf{Spearman p-value} & \textbf{Kendall p-value} & \textbf{Language} \\
        \midrule
        \dpskrone & $<10^{-115}$ & $<10^{-115}$ & $<10^{-57}$ & Chinese \\
        \gptfouro & $<10^{-109}$ & $<10^{-109}$ & $<10^{-56}$ & Chinese \\
        \dpskrone & $<10^{-237}$ & $<10^{-237}$ & $<10^{-126}$ & English \\
        \gptfouro & $<10^{-264}$ & $<10^{-264}$ & $<10^{-132}$ & English \\
        \bottomrule
    \end{tabular}
\end{table*}

\subsection{Detailed Evaluation Results for \dpskrone and \gptfouro}
\label{sec:detailed_result}
To illustrate the experimental results of different SDMs on the \ccc, evaluated separately by \dpskrone and \gptfouro, Table~\ref{tab:dpsk_benchmark_results} and Table~\ref{tab:gpt_benchmark_results} present the detailed results corresponding to those summarized in Table~\ref{tab:combined_results}.

\begin{table*}[htbp]
\centering
\tiny
\setlength{\tabcolsep}{2pt}
\caption{Accuracy (\%) of different SDMs on the Chinese (``zh'') or English (``en'') dialogue data subset of \ccc(\dpskrone).}
\label{tab:dpsk_benchmark_results}
\begin{tabular}{l *{11}{cc}}
\toprule
\multirow{2}{*}{\textbf{Category}}
& \multicolumn{2}{c}{\textbf{Freeze-Omni}}
& \multicolumn{2}{c}{\textbf{GLM-4-Voice}}
& \multicolumn{2}{c}{\textbf{\makecell{GPT-4o-\\Audio-Prev.}}}
& \multicolumn{2}{c}{\textbf{Kimi-Audio}}
& \multicolumn{2}{c}{\textbf{\makecell{LLaMA-\\Omni}}}
& \multicolumn{2}{c}{\textbf{\makecell{MooER-\\Omni}}}
& \multicolumn{2}{c}{\textbf{Moshi}}
& \multicolumn{2}{c}{\textbf{\makecell{Qwen2.5-\\Omni}}}
& \multicolumn{2}{c}{\textbf{Step-Audio}}
& \multicolumn{2}{c}{\textbf{VITA-Audio}}
& \multicolumn{2}{c}{\textbf{Overall}} \\
\cmidrule(lr){2-3} \cmidrule(lr){4-5} \cmidrule(lr){6-7} \cmidrule(lr){8-9} \cmidrule(lr){10-11} \cmidrule(lr){12-13} \cmidrule(lr){14-15} \cmidrule(lr){16-17} \cmidrule(lr){18-19} \cmidrule(lr){20-21} \cmidrule(lr){22-23} 
& \textbf{zh} & \textbf{en} & \textbf{zh} & \textbf{en} & \textbf{zh} & \textbf{en} & \textbf{zh} & \textbf{en} & \multicolumn{2}{c}{\textbf{en}} & \textbf{zh} & \textbf{en} & \multicolumn{2}{c}{\textbf{en}} & \textbf{zh} & \textbf{en} & \textbf{zh} & \textbf{en} & \textbf{zh} & \textbf{en} & \textbf{zh} & \textbf{en} \\
\midrule
Phonological & 16.22 & 6.90 & 18.92 & 20.69 & \textbf{29.73} & \textbf{44.83} & 18.92 & \textbf{44.83} & \multicolumn{2}{c}{17.24} & 18.92 & 20.69 & \multicolumn{2}{c}{10.34} & 27.03 & 37.93 & 21.62 & 27.59 & 8.11 & 27.59 & 19.93 & 25.86 \\
Semantic & 1.69 & 11.76 & 1.69 & 11.76 & 4.24 & \textbf{68.63} & 2.54 & 19.61 & \multicolumn{2}{c}{9.80} & 2.54 & 37.25 & \multicolumn{2}{c}{7.84} & \textbf{5.93} & 21.57 & \textbf{5.93} & 17.65 & 2.54 & 17.65 & 3.39 & 22.35 \\
\addlinespace
\textbf{\cambigdata} & 8.96 & 9.33 & 10.31 & 16.23 & \textbf{16.98} & \textbf{56.73} & 10.73 & 32.22 & \multicolumn{2}{c}{13.52} & 10.73 & 28.97 & \multicolumn{2}{c}{9.09} & 16.48 & 29.75 & 13.78 & 22.62 & 5.33 & 22.62 & 11.66 & 24.11 \\
\midrule
Omission & 4.29 & 7.84 & 4.29 & 6.86 & \textbf{45.71} & \textbf{16.67} & 27.14 & 12.75 & \multicolumn{2}{c}{6.86} & 32.86 & 7.84 & \multicolumn{2}{c}{4.90} & 25.71 & 14.71 & 17.14 & 11.76 & 5.71 & 7.84 & 20.36 & 9.80 \\
Coreference & 13.33 & 48.15 & 20.00 & 67.96 & \textbf{55.00} & \textbf{89.81} & 40.00 & 85.56 & \multicolumn{2}{c}{55.00} & 28.33 & 35.74 & \multicolumn{2}{c}{29.81} & 50.00 & 65.74 & 53.33 & 55.74 & 33.33 & 73.70 & 36.67 & 60.72 \\
Multi-turn & 7.89 & 32.35 & 10.53 & 58.82 & 10.53 & 47.06 & / & / & \multicolumn{2}{c}{47.06} & 60.53 & 38.24 & \multicolumn{2}{c}{/} & \textbf{84.21} & \textbf{97.06} & 5.26 & 32.35 & 52.63 & 52.94 & 33.08 & 50.74 \\
\addlinespace
\textbf{\ccontextdata} & 8.50 & 29.45 & 11.60 & 44.55 & 37.08 & 51.18 & 33.57 & 49.15 & \multicolumn{2}{c}{36.31} & 40.57 & 27.27 & \multicolumn{2}{c}{17.36} & \textbf{53.31} & \textbf{59.17} & 25.25 & 33.29 & 30.56 & 44.83 & 30.06 & 39.26 \\
\midrule
\textbf{Overall} & 8.68 & 21.40 & 11.09 & 33.22 & 29.04 & \textbf{53.40} & 22.15 & 40.68 & \multicolumn{2}{c}{27.19} & 28.64 & 27.95 & \multicolumn{2}{c}{13.23} & \textbf{38.58} & 47.40 & 20.66 & 29.02 & 20.47 & 35.94 & 22.41 & 32.94 \\
\bottomrule
\end{tabular}

\end{table*}

\begin{table*}[htbp]
\centering
\tiny
\setlength{\tabcolsep}{2pt}
\caption{Accuracy (\%) of different SDMs on the Chinese (``zh'') or English (``en'') dialogue data subset of \ccc(\gptfouro).}
\label{tab:gpt_benchmark_results}
\begin{tabular}{l *{11}{cc}}
\toprule
\multirow{2}{*}{\textbf{Category}}
& \multicolumn{2}{c}{\textbf{Freeze-Omni}}
& \multicolumn{2}{c}{\textbf{GLM-4-Voice}}
& \multicolumn{2}{c}{\textbf{\makecell{GPT-4o-\\Audio-Prev.}}}
& \multicolumn{2}{c}{\textbf{Kimi-Audio}}
& \multicolumn{2}{c}{\textbf{\makecell{LLaMA-\\Omni}}}
& \multicolumn{2}{c}{\textbf{\makecell{MooER-\\Omni}}}
& \multicolumn{2}{c}{\textbf{Moshi}}
& \multicolumn{2}{c}{\textbf{\makecell{Qwen2.5-\\Omni}}}
& \multicolumn{2}{c}{\textbf{Step-Audio}}
& \multicolumn{2}{c}{\textbf{VITA-Audio}}
& \multicolumn{2}{c}{\textbf{Overall}} \\
\cmidrule(lr){2-3} \cmidrule(lr){4-5} \cmidrule(lr){6-7} \cmidrule(lr){8-9} \cmidrule(lr){10-11} \cmidrule(lr){12-13} \cmidrule(lr){14-15} \cmidrule(lr){16-17} \cmidrule(lr){18-19} \cmidrule(lr){20-21} \cmidrule(lr){22-23} 
& \textbf{zh} & \textbf{en} & \textbf{zh} & \textbf{en} & \textbf{zh} & \textbf{en} & \textbf{zh} & \textbf{en} & \multicolumn{2}{c}{\textbf{en}} & \textbf{zh} & \textbf{en} & \multicolumn{2}{c}{\textbf{en}} & \textbf{zh} & \textbf{en} & \textbf{zh} & \textbf{en} & \textbf{zh} & \textbf{en} & \textbf{zh} & \textbf{en} \\
\midrule
Phonological & 16.22 & 10.34 & 18.92 & 34.48 & \textbf{29.73} & \textbf{62.07} & 21.62 & 48.28 & \multicolumn{2}{c}{13.79} & 21.62 & 17.24 & \multicolumn{2}{c}{10.34} & 27.03 & 58.62 & 24.32 & 31.03 & 8.11 & 34.48 & 20.95 & 32.07 \\
Semantic & 1.69 & 11.76 & 3.39 & 19.61 & \textbf{7.63} & \textbf{72.55} & 5.93 & 39.22 & \multicolumn{2}{c}{15.69} & 1.69 & 54.90 & \multicolumn{2}{c}{11.76} & \textbf{7.63} & 43.14 & 4.24 & 25.49 & 4.24 & 19.61 & 4.56 & 31.37 \\
\addlinespace
\textbf{\cambigdata} & 8.96 & 11.05 & 11.15 & 27.05 & \textbf{18.68} & \textbf{67.31} & 13.78 & 43.75 & \multicolumn{2}{c}{14.74} & 11.66 & 36.07 & \multicolumn{2}{c}{11.05} & 17.33 & 50.88 & 14.28 & 28.26 & 6.17 & 27.05 & 12.75 & 31.72 \\
\midrule
Omission & 4.29 & 5.88 & 7.14 & 5.88 & \textbf{42.86} & \textbf{15.69} & 31.43 & 7.84 & \multicolumn{2}{c}{4.90} & 31.43 & 1.96 & \multicolumn{2}{c}{0.98} & 30.00 & \textbf{15.69} & 18.57 & 9.80 & 7.14 & 7.84 & 21.61 & 7.65 \\
Coreference & 8.33 & 46.30 & 13.33 & 70.00 & 53.33 & \textbf{92.41} & 40.00 & 89.26 & \multicolumn{2}{c}{58.89} & 36.67 & 36.30 & \multicolumn{2}{c}{19.44} & \textbf{61.67} & 70.56 & 48.33 & 58.89 & 33.33 & 75.93 & 36.88 & 61.80 \\
Multi-turn & 15.79 & 55.88 & 10.53 & 58.82 & 15.79 & 47.06 & / & / & \multicolumn{2}{c}{64.71} & 65.79 & 44.12 & \multicolumn{2}{c}{/} & \textbf{81.58} & \textbf{94.12} & 10.53 & 50.00 & 73.68 & 67.65 & 39.10 & 60.29 \\
\addlinespace
\textbf{\ccontextdata} & 9.47 & 36.02 & 10.33 & 44.90 & 37.33 & 51.72 & 35.71 & 48.55 & \multicolumn{2}{c}{42.83} & 44.63 & 27.46 & \multicolumn{2}{c}{10.21} & \textbf{57.75} & \textbf{60.12} & 25.81 & 39.56 & 38.05 & 50.47 & 32.39 & 41.19 \\
\midrule
\textbf{Overall} & 9.26 & 26.03 & 10.66 & 37.76 & 29.87 & \textbf{57.95} & 24.75 & 46.15 & \multicolumn{2}{c}{31.60} & 31.44 & 30.90 & \multicolumn{2}{c}{10.63} & \textbf{41.58} & 56.42 & 21.20 & 35.04 & 25.30 & 41.10 & 24.26 & 37.36 \\
\bottomrule
\end{tabular}

\end{table*}

\begin{figure}[htbp]
    \centering
    \includegraphics[width=\linewidth]{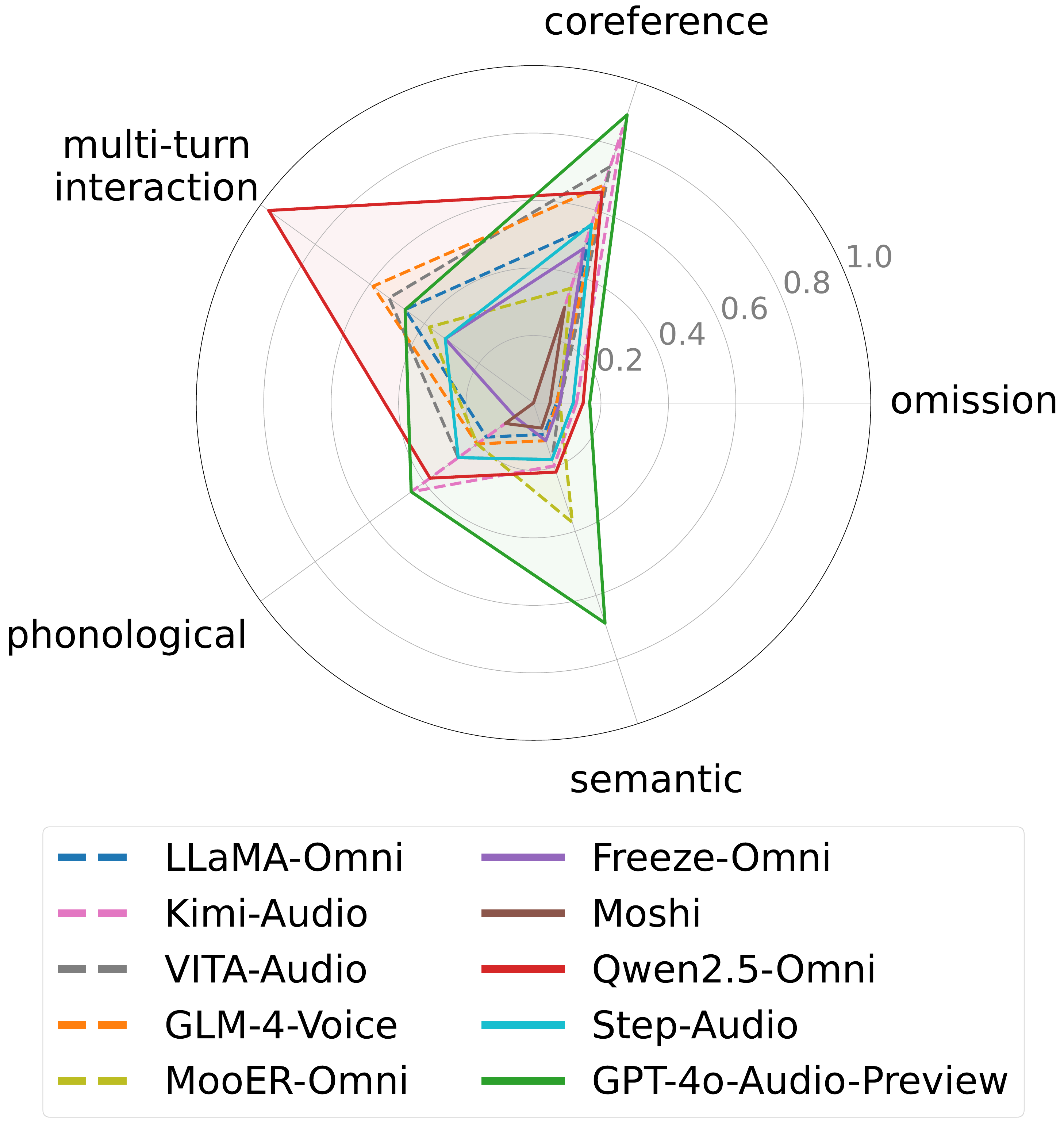}
    \caption{Radar charts depicting the experimental results of each SDM on the English portion of the dataset, assessed using \dpskrone.}
    \label{en-dpsk}
\end{figure}

\begin{figure}[htbp]
    \centering
    \includegraphics[width=\linewidth]{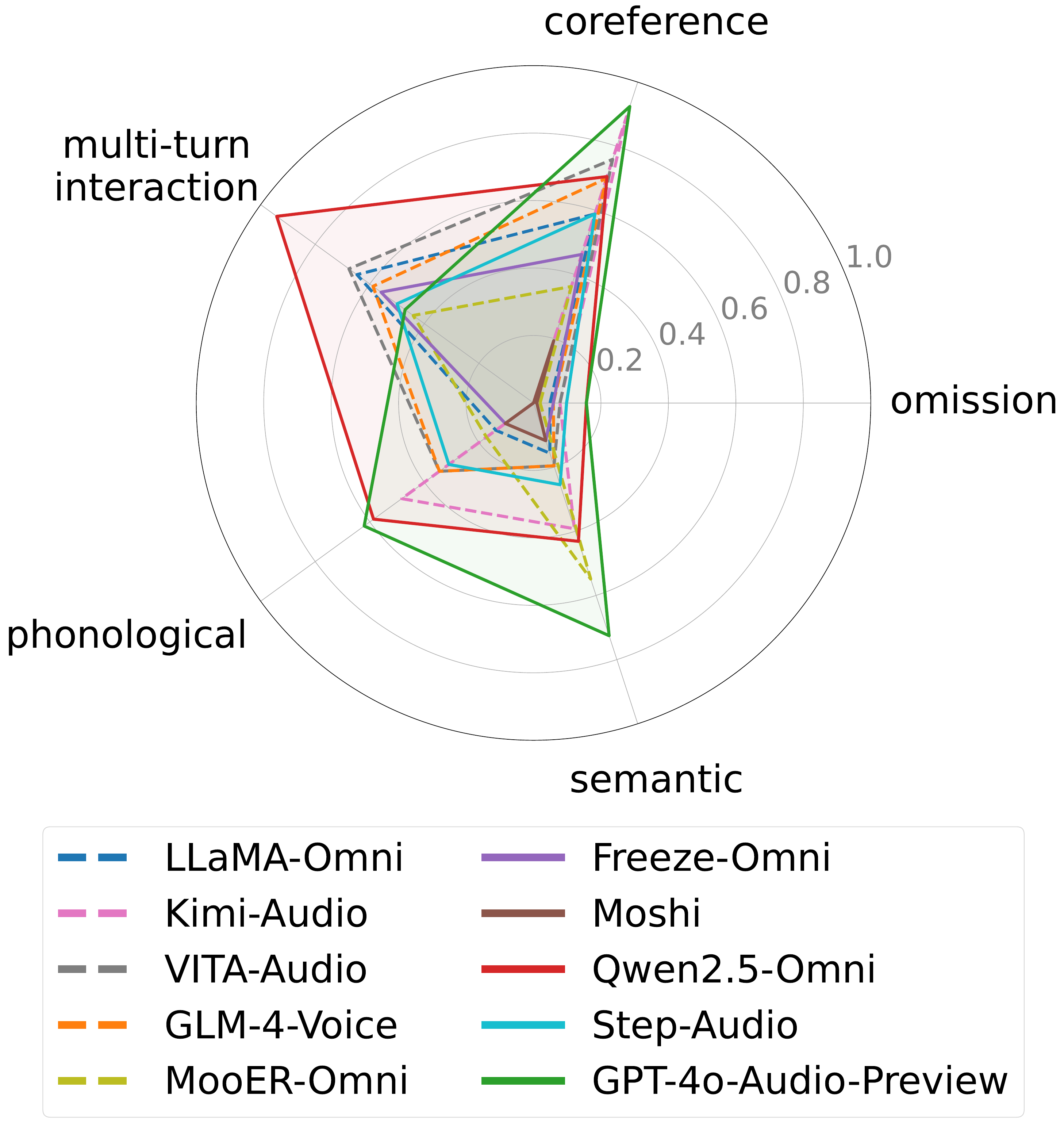}
    \caption{Radar charts depicting the experimental results of each SDM on the English portion of the dataset, assessed using \gptfouro.}
    \label{en-gpt}
\end{figure}

\begin{figure}[htbp]
    \centering
    \includegraphics[width=\linewidth]{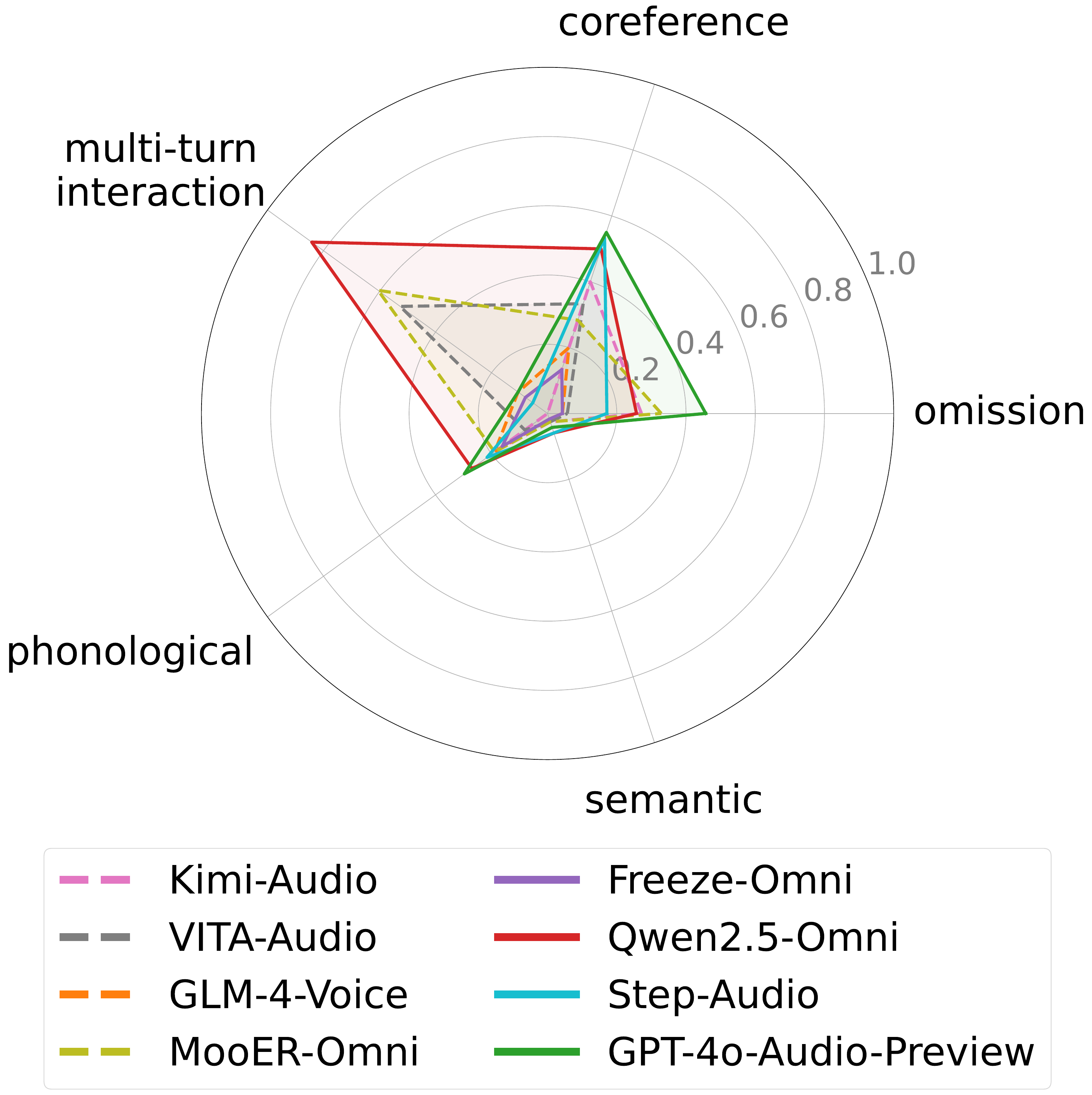}
    \caption{Radar charts depicting the experimental results of each SDM on the Chinese portion of the dataset, assessed using \dpskrone.}
    \label{cn-dpsk}
\end{figure}

\begin{figure}[htbp]
    \centering
    \includegraphics[width=\linewidth]{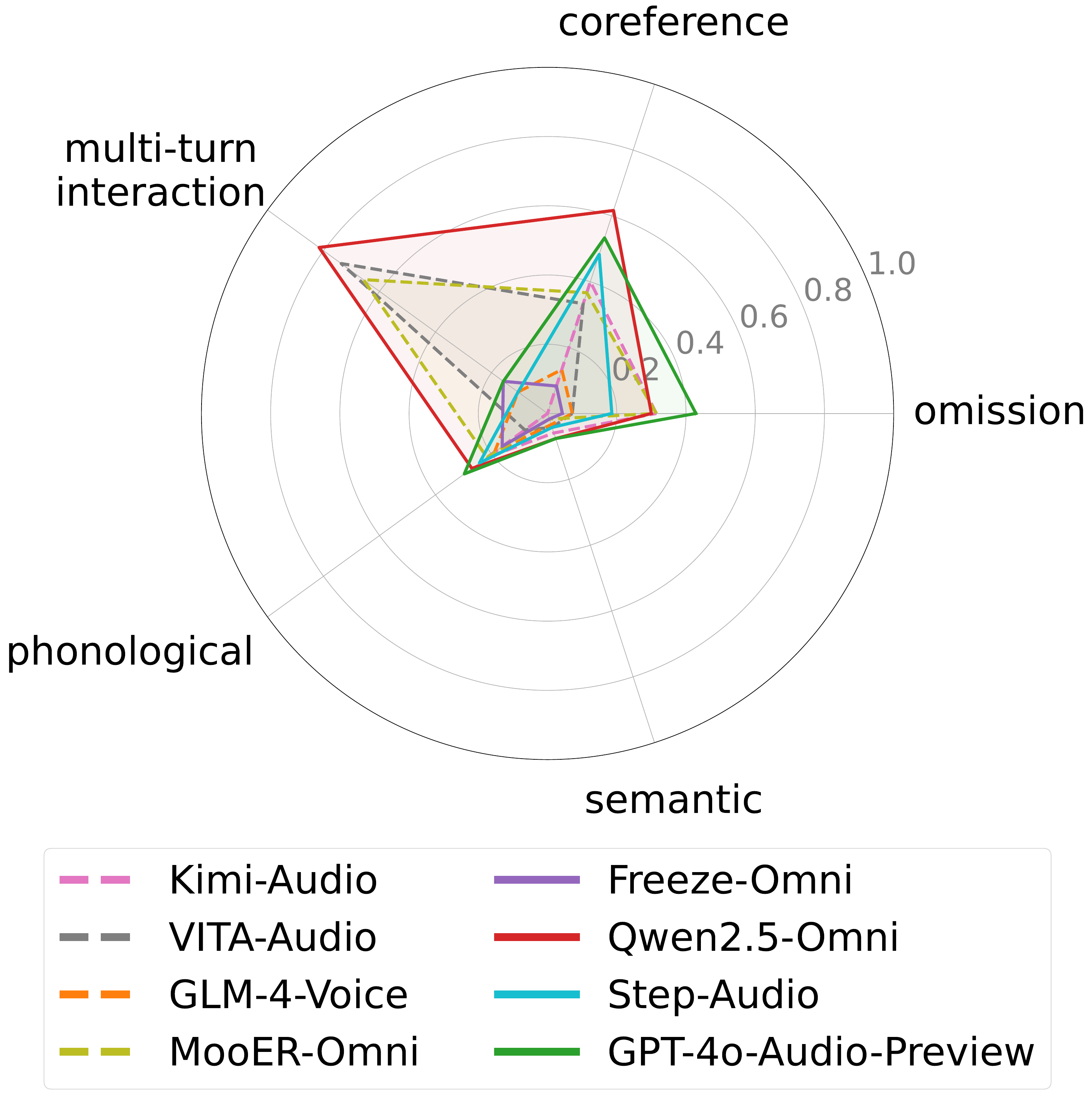}
    \caption{Radar charts depicting the experimental results of each SDM on the Chinese portion of the dataset, assessed using \gptfouro.}
    \label{cn-gpt}
\end{figure}

To present the radar charts of evaluation results for different SDMs on the Chinese and English sections of \ccc, evaluated respectively by \dpskrone and \gptfouro, we include 
Figure~\ref{en-dpsk} -~\ref{cn-gpt}
, which correspond to the summaries shown in Figure~\ref{en_gpt} and Figure~\ref{cn_gpt}.

To illustrate the experimental results of different SDMs on the omission and coreference sections, evaluated respectively by \dpskrone and \gptfouro, Table~\ref{tab:omi_and_cor_gpt} and Table~\ref{tab:omi_and_cor_dpsk} are presented, corresponding to the summary in Table~\ref{tab:merged_analysis}.

\begin{table*}[htbp]
    
    \scriptsize
    \centering
    \setlength{\tabcolsep}{3pt}
    \caption{Accuracy (\%) of omission and coreference phenomena (\gptfouro).}
    \label{tab:omi_and_cor_gpt}
    \begin{tabular}{ccccccccccccc}
    \toprule
    \multirow{1}{*}{\textbf{Phenomenon}} & \multirow{1}{*}{\textbf{Ability}} & \multirow{1}{*}{\textbf{Lang}}& \textbf{\makecell{Freeze-\\Omni}}& \textbf{\makecell{GLM-4-\\Voice}}& \textbf{\makecell{GPT-4o-\\Audio-Prev.}}& \textbf{\makecell{Kimi-\\Audio}}& \textbf{\makecell{LLaMA-\\Omni}}& \textbf{\makecell{MooER-\\Omni}}& \textbf{Moshi}& \textbf{\makecell{Qwen2.5-\\Omni}}& \textbf{\makecell{Step-\\Audio}}& \textbf{\makecell{VITA-\\Audio}} \\
    \midrule

    \multirow{4}{*}{Omission}
    & \multirow{2}{*}{Detection} & zh & 8.57 & 14.29 & \textbf{82.86} & 57.14 & / & 60.00 & / & 51.43 & 31.43 & 8.57 \\
    & & en & 7.84 & 3.92 & 11.76 & 7.84 & 5.88 & 1.96 & 1.96 & \textbf{13.73} & 5.88 & 7.84 \\
    \cmidrule(lr){2-13}
    & \multirow{2}{*}{Completion} & zh & 0.00 & 0.00 & 2.86 & 5.71 & / & 2.86 & / & \textbf{8.57} & 5.71 & 5.71 \\
    & & en & 3.92 & 7.84 & \textbf{19.61} & 7.84 & 3.92 & 1.96 & 0.00 & 17.65 & 13.73 & 7.84 \\
    \midrule

    \multirow{4}{*}{Coreference}
    & \multirow{2}{*}{Detection} & zh & 16.67 & 26.67 & 63.33 & 56.67 & / & 60.00 & / & \textbf{93.33} & 66.67 & 53.33 \\
    & & en & 52.22 & 84.44 & 95.93 & \textbf{98.52} & 78.89 & 24.44 & 25.56 & 71.11 & 65.56 & 87.78 \\
    \cmidrule(lr){2-13}
    & \multirow{2}{*}{Resolution} & zh & 0.00 & 0.00 & \textbf{43.33} & 23.33 & / & 13.33 & / & 30.00 & 30.00 & 13.33 \\
    & & en & 40.37 & 55.56 & \textbf{88.89} & 80.00 & 38.89 & 48.15 & 13.33 & 70.00 & 52.22 & 64.07 \\
    \bottomrule
    \end{tabular}
    
\end{table*}

\begin{table*}[htbp]
    
    \scriptsize
    \centering
    \setlength{\tabcolsep}{3pt}
    \caption{Accuracy (\%) of omission and coreference phenomena (\dpskrone).}
    \label{tab:omi_and_cor_dpsk}
    \begin{tabular}{ccccccccccccc}
    \toprule
    \multirow{1}{*}{\textbf{Phenomenon}} & \multirow{1}{*}{\textbf{Ability}} & \multirow{1}{*}{\textbf{Lang}}& \textbf{\makecell{Freeze-\\Omni}}& \textbf{\makecell{GLM-4-\\Voice}}& \textbf{\makecell{GPT-4o-\\Audio-Prev.}}& \textbf{\makecell{Kimi-\\Audio}}& \textbf{\makecell{LLaMA-\\Omni}}& \textbf{\makecell{MooER-\\Omni}}& \textbf{Moshi}& \textbf{\makecell{Qwen2.5-\\Omni}}& \textbf{\makecell{Step-\\Audio}}& \textbf{\makecell{VITA-\\Audio}} \\
    \midrule

    \multirow{4}{*}{Omission}
    & \multirow{2}{*}{Detection} & zh & 8.57 & 5.71 & \textbf{82.86} & 48.57 & / & 62.86 & / & 45.71 & 34.29 & 8.57 \\
    & & en & 9.80 & 5.88 & 15.69 & \textbf{17.65} & 7.84 & 11.76 & 5.88 & 9.80 & 9.80 & 5.88 \\
    \cmidrule(lr){2-13}
    & \multirow{2}{*}{Completion} & zh & 0.00 & 2.86 & \textbf{8.57} & 5.71 & / & 2.86 & / & 5.71 & 0.00 & 2.86 \\
    & & en & 5.88 & 7.84 & 17.65 & 7.84 & 5.88 & 3.92 & 3.92 & \textbf{19.61} & 13.73 & 9.80 \\
    \midrule

    \multirow{4}{*}{Coreference}
    & \multirow{2}{*}{Detection} & zh & 23.33 & 40.00 & 63.33 & 63.33 & / & 56.67 & / & \textbf{80.00} & 73.33 & 60.00 \\
    & & en & 62.96 & 83.33 & 94.81 & \textbf{95.56} & 78.15 & 27.41 & 45.19 & 70.00 & 61.11 & 87.41 \\
    \cmidrule(lr){2-13}
    & \multirow{2}{*}{Resolution} & zh & 3.33 & 0.00 & \textbf{46.67} & 16.67 & / & 0.00 & / & 20.00 & 33.33 & 6.67 \\
    & & en & 33.33 & 52.59 & \textbf{84.81} & 75.56 & 31.85 & 44.07 & 14.44 & 61.48 & 50.37 & 60.00 \\
    \bottomrule
    \end{tabular}
    
\end{table*}

\end{document}

%% file: tool.tex
\usepackage{enumitem}
\usepackage{verbatim}
\usepackage{xspace}
\usepackage{diagbox}
\usepackage{array}
\usepackage{cite}

\usepackage{multirow}
\usepackage{diagbox}
\usepackage{url}
\usepackage{comment}
\usepackage{booktabs}
\usepackage{amsmath}
\usepackage{scalerel}
\usepackage[figuresright]{rotating}
\usepackage{threeparttable}
\usepackage{algorithm}
\usepackage{algorithmic}
\usepackage{siunitx}
\usepackage{stackengine}
\usepackage{pifont}

\usepackage{booktabs, makecell, multirow}
\usepackage{siunitx} 

\usepackage{tcolorbox}
\definecolor{grey}{RGB}{128,128,128}

\newcommand{\ccc}{C$^3$\xspace}
\newcommand{\cdata}{C\textsubscript{data}\xspace}
\newcommand{\cambigdata}{C\textsubscript{am-data}\xspace}
\newcommand{\ccontextdata}{C\textsubscript{con-data}\xspace}
\newcommand{\gptfouroa}{GPT-4o-Audio-Preview\xspace}
\newcommand{\gptfouro}{GPT-4o\xspace}
\newcommand{\dpskrone}{DeepSeek-R1\xspace}

\usepackage{mathtools}
\usepackage{multirow}
\usepackage{multicol}
\usepackage{booktabs}
\usepackage{subfigure}

\usepackage{amsmath,amsfonts,bm}

\def\eqref#1{equation~\ref{#1}}

\def\1{\bm{1}}

\DeclareMathAlphabet{\mathsfit}{\encodingdefault}{\sfdefault}{m}{sl}
\SetMathAlphabet{\mathsfit}{bold}{\encodingdefault}{\sfdefault}{bx}{n}

